\documentclass[sigconf]{aamas} 

\usepackage{caption}
\usepackage{subcaption}
\usepackage{graphicx}
\usepackage{mathtools}
\usepackage{placeins}
\usepackage{balance} 

\setcopyright{ifaamas}
\acmConference[AAMAS '22]{Proc.\@ of the 21st International Conference
on Autonomous Agents and Multiagent Systems (AAMAS 2022)}{May 9--13, 2022}
{Online}{P.~Faliszewski, V.~Mascardi, C.~Pelachaud,
M.E.~Taylor (eds.)}
\copyrightyear{2022}
\acmYear{2022}
\acmDOI{}
\acmPrice{}
\acmISBN{}

\acmSubmissionID{572}

\title{Scalable Multi-Agent Model-Based Reinforcement Learning}

\author{Vladimir Egorov}
\affiliation{
  \institution{JetBrains Research; HSE University}
  \city{Saint Petersburg}
  \country{Russian Federation}}
\email{vsegorov@edu.hse.ru}

\author{Alexei Shpilman}
\affiliation{
  \institution{JetBrains Research; HSE University}
  \city{Saint Petersburg}
  \country{Russian Federation}}
\email{alexey@shpilman.com}

\begin{abstract}
Recent Multi-Agent Reinforcement Learning (MARL) literature has been largely focused on Centralized Training with Decentralized Execution (CTDE) paradigm. CTDE has been a dominant approach for both cooperative and mixed environments due to its capability to efficiently train decentralized policies. While in mixed environments full autonomy of the agents can be a desirable outcome, cooperative environments allow agents to share information to facilitate coordination. Approaches that leverage this technique are usually referred as communication methods, as full autonomy of agents is compromised for better performance. Although communication approaches have shown impressive results, they do not fully leverage this additional information during training phase. In this paper, we propose a new method called MAMBA which utilizes Model-Based Reinforcement Learning (MBRL) to further leverage centralized training in cooperative environments. We argue that communication between agents is enough to sustain a world model for each agent during execution phase while imaginary rollouts can be used for training, removing the necessity to interact with the environment. These properties yield sample efficient algorithm that can scale gracefully with the number of agents. We empirically confirm that MAMBA achieves good performance while reducing the number of interactions with the environment up to an orders of magnitude compared to Model-Free state-of-the-art approaches in challenging domains of SMAC and Flatland. \footnote{The code for this paper can be found at \href{https://github.com/jbr-ai-labs/mamba}{\color{blue}{https://github.com/jbr-ai-labs/mamba}}}
\end{abstract}

\keywords{Multi-Agent Reinforcement Learning; Model-Based Reinforcement Learning; Communication}

\newcommand{\BibTeX}{\rm B\kern-.05em{\sc i\kern-.025em b}\kern-.08em\TeX}

\begin{document}


\pagestyle{fancy}
\fancyhead{}


\maketitle 


\section{Introduction}

Reinforcement Learning (RL) has achieved numerous breakthroughs \cite{berner2019dota, vinyals2019grandmaster, alphago} in recent years both for single- and multi-agent environments. However, sample efficiency has always been a prominent issue as a tremendous number of interactions is required by Model-Free Reinforcement Learning (MFRL) approaches to achieve optimal performance. Model-Based Reinforcement Learning (MBRL) tries to alleviate this issue by allowing the agent to interact with a learned model of the environment instead. It has shown success both in environments with known world dynamics such as shogi or go \cite{schrittwieser2020mastering} and where dynamics need to be learned from samples such as Atari \cite{kaiser2019model, hafner2020mastering}. However, translating this success to multi-agent settings is non-trivial. Several studies \cite{krupnik2020multi, schwarting2021deep} have shown that MBRL approaches can be much more sample efficient than their MFRL counterparts in environments with two agents. Nonetheless, all of the current state-of-the-art MFRL approaches on popular multi-agent benchmarks such as StarCraft Multi-Agent Challenge \cite{samvelyan19smac} remain uncontested.


One of the major paradigms in Multi-Agent Reinforcement Learning is Centralized Training with Decentralized Execution (CTDE) \cite{foerster2018counterfactual, rashid2018qmix, lowe2017multi}, which holds to the premise that we can use additional information during training as long as execution remains decentralized. This paradigm is well-motivated when we want full autonomy for the agents \cite{maes1993modeling}, for example, when the task is not fully cooperative because agents might not want to share their information with others \cite{cao2018emergent, blumenkamp2020emergence,noukhovitch2021emergent}. Furthermore, in many real world scenarios, such as autonomous driving \cite{desjardins2011cooperative} or drone management \cite{azar2021drone}, agents must collaborate with each other to achieve their goals and they can also share information for better coordination. This setup is usually defined as a communication technique \cite{foerster2016learning, liu2020multi}, as the agents no longer have full autonomy yet they can leverage the messages from others to make their decisions. As a part of CTDE, communication methods usually restrict the message channel either by limiting bandwidth or receiving messages only from the neighbours in the environment \cite{gronauer2021multi}. Other CTDE techniques can still be applied together with the communication technique to stabilize learning or solve the credit assignment issue. Note that if we decide to centralize execution even more, reducing the multi-agent problem to single-agent, this solution will suffer from scalability issues as the number of states and actions will grow exponentially with the number of agents. Thus, communication methods strike a balance between autonomy of the agents and efficiently using available information to make decisions in collaborative environments. However, previous studies do not fully leverage this additional shared information during the training phase. 


In this paper, we show that communication can be sufficient to sustain a world model, making it possible to exploit centralized training further by training policies without interacting with the environment. Furthermore, it is sufficient to only communicate discrete messages with the neighbouring agents to update a world model during execution phase. We denote our approach as MAMBA, which is derived from Multi-Agent Model-Based Approach. To our knowledge, this is the first method that extends world models to environments with an arbitrary number of agents and that views world models as an instance of communication. We show that our method can reduce the required number of samples by an order of magnitude compared to MFRL approaches and consistently outperform current state-of-the-art methods in low data regime in challenging environments of SMAC \cite{samvelyan19smac} and Flatland \cite{mohanty2020flatland}.

\section{Definitions and Background}

\subsection{Markov Game}


A Markov game \cite{littman1994markov} is a tuple $\mathcal{G} = (S, n, \mathcal{A}, O, T, r)$, where $S$ is a set of states $s$, $n$ is a number of players, $\mathcal{A}_i$ is a set of actions $a$ of player $i$. $\mathcal{O}: S \times N \rightarrow \mathbb{R}^d$ is an observation function that specifies d-dimensional observations available to each agent and $O_i$ is set of observations $o_i$ of agent $i$. $T: S \times \mathcal{A}_1 \times \dots \times \mathcal{A}_n \rightarrow \Delta(S)$ and $r_i: S \times \mathcal{A}_1 \times \dots \times \mathcal{A}_n \rightarrow \mathbb{R}$ are transition and reward functions respectively, where $\Delta$ is a discrete probability distribution over $S$. We define $\Delta_0(S)$ as the distribution of initial states and return of player $i$ in state $s$ with discount factor $\gamma \in [0, 1)$ as $R_i(s) = \overset{\infty}{\underset{t=0}{\sum}} \left[ \gamma^t r_i(s_t, a^1_{t}, \dots, a^n_{t}) \mid s_0 = s \right]$. Each player has its own policy $\pi_i: O_i \rightarrow \Delta(\mathcal{A}_i)$ that returns probability for each action to be taken when observing $o_i$ and value function  $V_{i}(s) = \mathop{\mathbb{E}}_{\pi_i}[R_i(s)]$. We also define Q-function as  $Q_{i}(s,a) = \mathop{\mathbb{E}}_{\pi_i}[R_i(s) \mid a^i_{0} = a]$. Bold font will denote vector, e.g. $\textbf{a} = (a^1, \dots, a^N)$ and subscript $a_t$ will denote current step in the environment.


\subsection{Single-Agent Reinforcement Learning}\label{sec-background-rl}

\paragraph{Model-Free Reinforcement Learning.}

There are two major MFRL frameworks, namely Q-learning \cite{watkins1992q} and Actor-Critic \cite{A3C}. In Actor-Critic framework, the Actor's goal is to learn a policy $\pi(s)$ that maximizes agents returns predicted by the Critic. A widely-used method is proximal policy optimization (PPO) \cite{schulman2017proximal}, where the Actor's neural network is trained on the following loss:

\begin{equation}\label{eq:loss_ppo}
\mathcal{L}_\pi(A_t) = -\min(\mathcal{R}_t A_t,  clip(\mathcal{R}_t, 1 - \epsilon, 1 + \epsilon) A_t)
\end{equation}

where  $\mathcal{R}_t = \frac{\pi(s_t)}{\pi_{old}(s_t)}$ denotes probability ratio of the policies after and before the update and $A_t$ is an advantage function predicted by the Critic. Using this loss ensures that the agent's policy makes a conservative update within a trust region. The advantage is defined as $A_t = y_t - V_t$, where $y_t = r_t + \gamma V_{t+1}$. The Critic's neural network is independently trained to predict $V_t$ by minimizing squared TD error $(y_t - V_t)^2$. In Q-learning framework, we do not have a policy $\pi(s)$ and instead use learned Q-function directly to predict next action.

\paragraph{Model-Based Reinforcement Learning.}

MBRL approaches \cite{sutton1991dyna} additionally learn a world model that includes both transition function $T$ and reward function $r$. Decoupling learning dynamics of the system from the decision making allows us to use more sophisticated unsupervised learning techniques to represent current state in latent space as well as to use imaginary rollouts for training \cite{kaiser2019model, hafner2020mastering}. Many previous works provide both theoretical and empirical evidence that MBRL approaches can be much more sample efficient than their MFRL counterparts. Recent breakthroughs of Dreamer \cite{hafner2019dream, hafner2020mastering} also showed that MBRL is capable of achieving same or better performance than Model-Free approaches both in continuous and discrete environments.

\subsection{Multi-Agent Reinforcement Learning}\label{sec-background-marl}


In Multi-Agent Reinforcement Learning (MARL), multiple agents learn and interact in the same environment. In this paper, we will focus on cooperative environments \cite{samvelyan19smac}, where agents have the same goal and they need to collaborate with each other to achieve it. One naive approach in MARL is training agents independently using unmodified single-agent RL techniques \cite{tan1993multi, de2020independent}. However, it does not have any convergence guaranties due to the inherent non-stationarity of multi-agent environments resulting from the constantly changing policies of other agents \cite{laurent2011world, non-stationarity}. On the other side of the spectrum, we could look at the multi-agent problem as a single-agent one. This yields a fully centralized approach that controls all agents simultaneously based on global information. Consequently, a centralized solution cannot scale to a large number of agents due to the resultant exponential growth of the action space, known as the familiar 'curse of dimensionality' \cite{sunehag2017value}. This leaves us with one viable technique to train agents in multi-agent environments, namely Centralized Training with Decentralized Execution. We also explicitly highlight the communication technique, which is considered part of CTDE \cite{gronauer2021multi}, that allows an additional message channel during execution for better coordination.


\paragraph{Centralized Training with Decentralized Execution}\label{par:comm}

CTDE is a combination of independent and centralized MARL \cite{kraemer2016multi,lowe2017multi, sunehag2017value, rashid2018qmix,foerster2018counterfactual,son2019qtran}. This paradigm allows additional information to be used in training provided that execution is still decentralized. This enhancement can help to alleviate the issues of credit assignment and non-stationarity as we no longer need to rely only on local observations and rewards during training. Decentralized execution also ensures that we deal with the curse of dimensionality as agents still make their decisions independently based on relevant local information.

Communication can be seen as a part of the CTDE paradigm as we can still use additional information on the training stage, but now we allow agents to communicate with each other during execution, which is a natural assumption for cooperative environments \cite{liu2020multi, sukhbaatar2016learning, foerster2016learning} where agents have the same goal and can also be used in mixed environments \cite{singh2018learning} to promote cooperation. Despite being considered CTDE \cite{gronauer2021multi}, we argue that communication can greatly benefit from possessing certain properties to ensure decentralized execution. In this paper, we will focus on the two desirable properties of communication that can greatly improve scalability and decentralization of the agents. The first property is called \emph{discrete communication} and is strongly connected with the limitation of the message channel bandwidth as it restricts agents to sending each other only discrete messages \cite{gronauer2021multi, foerster2016learning}. This property allows us to significantly reduce the number of bits for message transmission and can be considered a universal restriction between the environments on the number of bits allowed for one message. The second property is called \emph{locality} \cite{zhang2018fully}, which can be naturally defined in most environments. This property allows an agent to communicate only with its neighbours on the map, which makes the decision making decentralized since each agent takes action individually based only on local information \cite{zhang2018fully}. Unfortunately, some environments do not exhibit this locality property either because the map is too small or coordination between all agents is required to solve the task.

\paragraph{Multi-Agent Deep Deterministic Policy Gradient (MADDPG)} \cite{lowe2017multi} MADDPG is a CTDE algorithm specifically designed for mixed environments. MADDPG shows that augmenting critics with the states and actions of all agents can reduce the variance of policy gradients, which can improve performance.

\paragraph{Counterfactual Multi-Agent policy gradients (COMA)} \cite{foerster2018counterfactual} COMA is an Actor-Critic algorithm designed to alleviate the credit assignment issue in multi-agent environments. The core idea is based on Difference Rewards \cite{wolpert2002optimal}, which compares the current global reward of the agent to the average reward that the agent can get in the current state. COMA applies this idea to Q-functions of agents, additionally using augmented critics similar to MADDPG.

\paragraph{QMIX} \cite{rashid2018qmix} QMIX is an algorithm designed for training Q-learning agents in cooperative environments. As with COMA, it tries to solve credit assignment using additional information during training. As agents still need the Q-function during execution, we cannot alter its input as in COMA. QMIX proposes to use a mixer function to combine Q-functions of agents into a centralized Q-function that is trained on global reward. QMIX ascertains that mixer function ensures monotonicity in its inputs, which makes the argmax of the agents' Q-functions consistent with the argmax of the centralized Q-function. This property is called Individual-Global-Max (IGM) and is essential for factorizing the global Q-function to agents' Q-Functions. 

\paragraph{QTRAN} \cite{son2019qtran} QTRAN is an algorithm that extends the class of functions that we can factorize compared to QMIX. QTRAN shows that the monotonicity property can be lifted from the mixer function and introduces additional "track" losses that keep the agents' Q-functions close to the global Q-function, which is sufficient to preserve the IGM property but introduces additional training complexity. 

\paragraph{CommNet} \cite{sukhbaatar2016learning} CommNet is a communication algorithm that employs continuous protocols. CommNet uses LSTM \cite{hochreiter1997long} to process messages and shows that communication can be essential in solving tasks in partially observable environments.

\paragraph{G2A} \cite{liu2020multi} G2A is an algorithm that uses two-stage attention architecture to process messages from other agents. The first stage uses Hard Attention to detect relevant agents in the environment, while the second stage uses Soft Attention to process messages from these agents.

\subsection{Attention Mechanism} 

Attention \cite{vaswani2017attention, bahdanau2014neural} has become an ubiquitous approach for sequence data processing, achieving state-of-the-art performance in multitude of fields such as NLP \cite{devlin2018bert} and CV \cite{dosovitskiy2020image}. It has also shown to be useful in MARL when processing input observations of all agents \cite{vinyals2019grandmaster} or solving credit assignment \cite{iqbal2019actor}.

In this paper, we will focus mainly on self-attention, which takes as an input value matrix $V$ with dimensionality $n \times d$ and for each of $n$ vectors outputs weighted average of the whole vector sequence. There are two types of attention mechanism, namely soft attention and hard attention \cite{shen2018reinforced}. Following \cite{vaswani2017attention}, the former assigns the weights for each value vector using this formula:

\begin{equation}\label{eq:attn}
    \text{Soft Self-Attention}(Q, K, V) = \text{softmax}(\frac{QK^T}{\sqrt{d}}) V
\end{equation}

\noindent where $Q$, $K$, $V$ are linear projections of the value vectors, and latter assigns either zeroes or ones depending on the relevance of the vector.

\section{MAMBA}

In this section, we describe our method called Multi-Agent Model-Based Approach (MAMBA). First, we formulate the desiderata for the method and review existing papers that tackle the outlined problems. Second, we describe our method and its communication properties.

\subsection{Desiderata}

\paragraph{Sample Efficiency} One of the main strengths of MBRL approaches is the sample efficiency that results from using imaginary rollouts provided by the learned model. In the single-agent setting, multiple papers provide empirical evidence \cite{kaiser2019model, hafner2019dream, hafner2020mastering} as well as theoretical justifications \cite{agarwal2020model, azar2013minimax, li2020breaking} that MBRL approaches achieve near-optimal sample efficiency compared to Model Free approaches. This theory is further extended to the multi-agent setting \cite{liu2021sharp, zhang2020model}. However, empirical performance is still lacking due to the increased complexity of the model and exponential growth of the action and state spaces.

\paragraph{Addressing non-stationarity} Experience replay buffer is a popular technique for off-policy learning that has been shown to improve sample efficiency compared to on-policy algorithms \cite{andrychowicz2017hindsight}. However, in multi-agent environments collected experience quickly becomes irrelevant due to the inherent non-stationarity \cite{non-stationarity}. Methods have been proposed by \cite{foerster2017stabilising} to alleviate non-stationarity using techniques such as importance sampling and fingerprinting. However, these techniques can suffer from large variance as the agents progressively change their policies \cite{precup2000eligibility}. Another approach is to use on-policy algorithms and a world model that approximates the environment's dynamics, which can be used to roll out imaginary trajectories starting from the states in the buffer for current policies, thus mitigating any issues with irrelevant experience \cite{zhang2020model}.

\paragraph{Scalability} One of the key problems of learning dynamics of multi-agent environments is the curse of dimensionality. We want the world model to scale to a large number of agents in the environment, which would require the model to manage the resultant exponential growth of state and action spaces. This would allow it to process large numbers of transitions during both training and inference phases.

\paragraph{Locality} Multiple studies \cite{wang2020few, liu2020multi} have shown that in large environments it is possible to significantly reduce the state space to comprise only relevant information for the agents' decision making. This reduction can be naturally applied to environments like autonomous driving \cite{singh2018learning}, where an agent only needs to be informed about cars in its local view. This property of locality can also be applied to communication \cite{liu2020multi, zhang2018fully} by the same reasoning. Ideally, a world model should have a locality property, so that if the environment allows us to define neighbours of the agent, we want to communicate only with them during execution. 

\paragraph{Discrete Communication} One dichotomy between the communication protocols is that the message can be either continuous \cite{sukhbaatar2016learning} or discrete \cite{foerster2016learning}. It is desirable that agents communicate using discrete messages, as it is more suitable for channels with limited bandwidth. Discrete representation allows us to greatly reduce the amount of information broadcast by the agents, which can be essential for deploying communication methods in the real world. Note that even with a restricted message channel, discrete messages can greatly enhance the performance of the agents \cite{foerster2016learning}.

\paragraph{Decentralized Execution} Finally, we want to deploy agents in a decentralized fashion, so they can make their decisions independently based on their local observations and messages from other agents \cite{zhang2018fully}. To this end, one of the main paradigms is CTDE \cite{rashid2018qmix, foerster2018counterfactual, liu2020multi, sukhbaatar2016learning}, which achieves strong performance in multi-agent domains while maintaining decentralization on the execution stage. We want to learn a world model that can be sustained in a decentralized manner during execution using only communication between agents.

\subsection{Architecture}

\begin{figure}[ht]
\centering
\centering
\includegraphics[width=\linewidth]{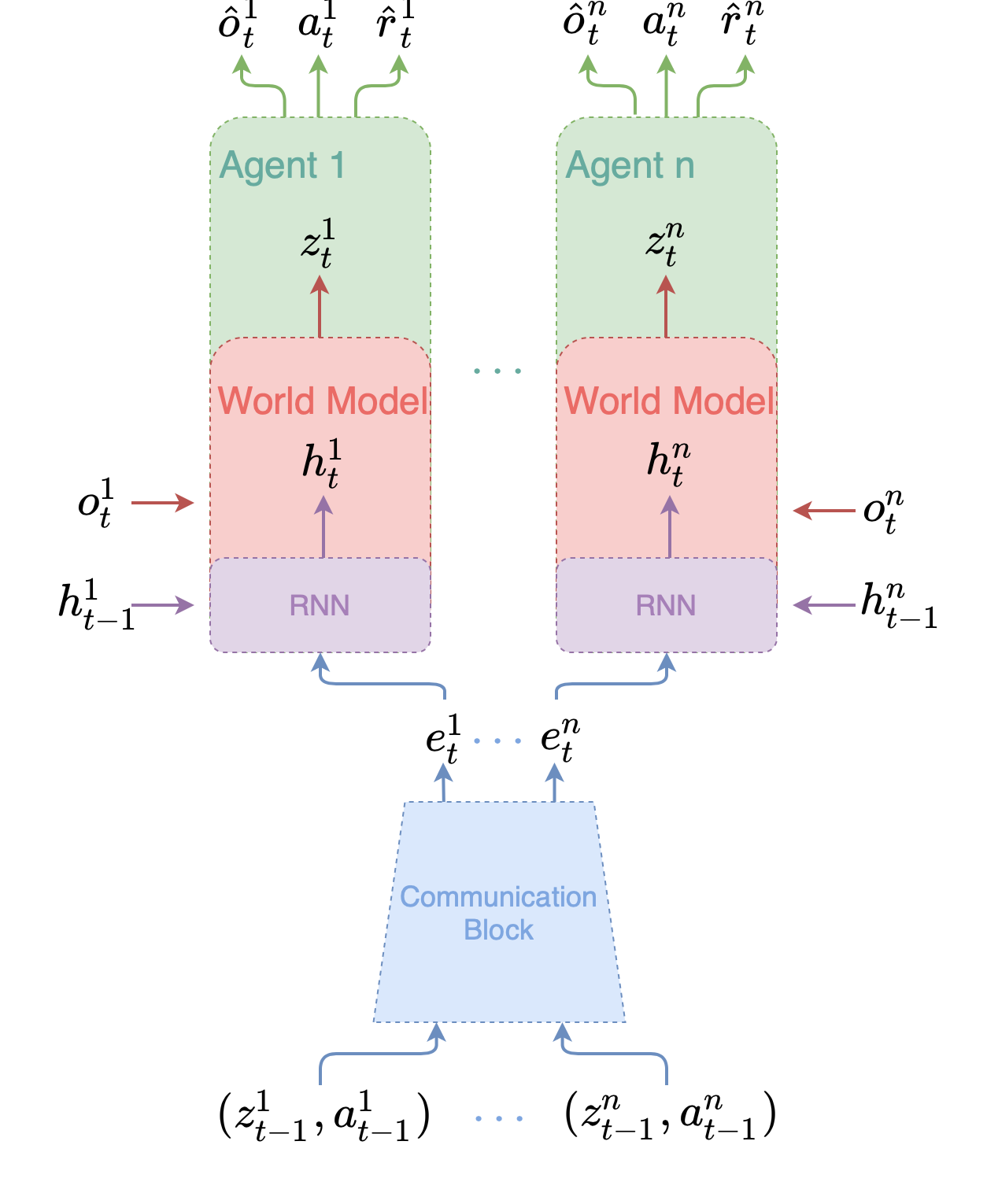}
\caption{Training Phase. Transitions $(o^i_{t-1}, a^i_{t-1}, o^i_{t})$ are sampled from the buffer. Agents use the Communication Block to predict their current state feature vectors $e^i_t$ from stochastic states and actions from previous step $(z^i_{t-1}, a^i_{t-1})$. These feature vectors are then used to update the hidden state of the model $h^i_t$. Agent then can predict current stochastic state $z^i_t$ from $h^i_t$ and its current observation $o^i_t$ and in turn use it to predict local information such as observation $\hat{o}^i_t$ and reward $\hat{r}^i_t$. Agent also uses stochastic state to predict its next action $a^i_t$. During the imaginary rollout stage, Communication Block utilizes predicted stochastic states $\hat{z}^i_t$ instead of $z^i_t$ and observations $o^i_t$ are not used.}
\label{fig:scheme_train}
\end{figure}

We chose DreamerV2 \cite{hafner2020mastering}, a state-of-the-art MBRL approach for discrete environments, as a backbone for our algorithm. Following Dreamer notation, the model consists of six components: 
\begin{align*}
&\text{RSSM}=\left\{\begin{array}{ll}  
    \text{Recurrent model:} \hspace{0.05cm}  &h^i_t = f_\phi(h^i_{t-1}, \textbf{z}_{t-1}, \textbf{a}_{t-1})\\[2pt]
   \text{Representation model:} \hspace{0.05cm}&z^i_t \sim q_\phi(z^i_t \mid h^i_t, o^i_t)\\[2pt]
   \text{Transition predictor:}\hspace{0.05cm}  &\hat{z}^i_t \sim p_\phi(\hat{z}^i_t \mid h^i_t)
\end{array}\right. \\
&\begin{array}{ll}
  \hspace{1.35cm} \text{Observation predictor:}        &\hat{o}^i_t \sim p_\phi(\hat{o}^i_t \mid h^i_t, z^i_t)\\[2pt]
  \hspace{1.35cm} \text{Reward predictor:}      &\hat{r}^i_t \sim p_\phi(\hat{r}^i_t \mid h^i_t, z^i_t)\\[2pt]
  \hspace{1.35cm} \text{Discount predictor:}    &\hat{\gamma}^i_t \sim p_\phi(\hat{\gamma}^i_t \mid h^i_t, z^i_t)
\end{array}
\end{align*}
where the RSSM model \cite{hafner2019learning} is used to learn the dynamics of the environment and  Observation,  Reward and Discount predictors are used for reconstructing corresponding variables.
We learn Observation, Reward and Discount predictors similar to \cite{hafner2020mastering} via supervised loss and Representation model by minimizing the evidence lower bound \cite{kingma2013auto}. Transition predictor is then trained by minimizing KL divergence between $\hat{z}^i_t$ and $z^i_t$, where $z^i_t$ is a stochastic state sampled from categorical distribution same as in DreamerV2. Recurrent model consists of 1 layer GRU \cite{cho2014learning} as in the original DreamerV2 implementation. In order to facilitate independence between hidden states $\textbf{h}_t$, we predict individual reward, discount and observation only from agent's latent state. We also use the sum of KL divergences between individual stochastic states $\hat{z}^i_t$ by applying the chain rule to the true Transition function:
\begin{equation}
p_\phi(\hat{\textbf{z}}_t \mid \textbf{h}_t) = 
p_\phi(\hat{z}^1_t \mid \textbf{h}_t) \cdots p_\phi(\hat{z}^n_t \mid \textbf{h}_t, \hat{z}^1_t, \ldots, \hat{z}^{n-1}_t)
\end{equation}
However, predicting latent states autoregressively is too impractical and we elaborate on its substitute in the next paragraph. The final loss for learning the world model consists of:
\begin{align*}
\begin{array}{ll} 
\text{Observation loss:} & -\ln{p_\phi(o^i_t \mid h^i_t, z^i_t)}\\[2pt]
\text{Reward loss:} &-\ln{p_\phi(\hat{r}^i_t \mid h^i_t, z^i_t)} \\[2pt]
\text{Discount loss:} & - \ln{p_\phi(\hat{\gamma}^i_t \mid h^i_t, z^i_t)}\\[2pt]
\text{KL divergence loss:} &\text{KL}\left[q_\phi(z^i_t \mid h^i_t, o^i_t) \lVert p_\phi(\hat{z}^i_t \mid h^i_t) \right]\\
\end{array}
\end{align*}
which is calculated in expectation over $q_\phi(\textbf{z}_{1:T} \mid \textbf{a}_{1:T}, \textbf{o}_{1:T}) $ and summed over all $n$ agents and rollout length $T$. As in Dreamer, all losses are optimized jointly using the Adam optimizer \cite{kingma2014adam}. Note that we do not use an additional $\beta$ weight for KL divergence \cite{higgins2016beta} to simplify algorithm complexity. As in DreamerV2, we use KL balancing to facilitate learning of prior $p_\phi(\hat{z}^i_t \mid h^i_t)$ by assigning a larger weight to cross entropy and a lower weight to posterior entropy in the KL term. The schematic representation of the training phase is in Figure \ref{fig:scheme_train}.

\paragraph{Transition function.} In our method, we implement Transition function using Attention. Simply predicting next latent states independently using Attention can create inconsistencies between agents' current views of the environment, which can hinder performance. To solve this issue, we could theoretically predict next latent states autoregressively, that is, condition the latent state of $i$-th agent on the predicted latent states of the previous agents. However, this would significantly slow down training and would not allow decentralized decision making during execution. Another approach, proposed by \cite{krupnik2020multi}, is to maximize mutual information between the latent state and the previous action of the agent:
\begin{equation}
    \mathcal{L}_{\text{info}} = -I((h^i_t, z^i_{t}); a^i_{t-1})
\end{equation} 
This way, we encourage the latent state to depend mostly on its own actions, thus making it less dependent on others. Consequently, this disentanglement of the agent's latent space allows us to dispatch them separately. We use the lower bound proposed by \cite{chen2016infogan} to practically maximize mutual information by training an additional neural network to predict previous action from the latent state of the agent:
\begin{equation}
    \mathcal{L}_{\text{info}} = -\ln{p_\phi(a^i_{t-1} \mid h^i_t, z^i_t)}
\end{equation} 

\paragraph{Stochastic state} During the execution and imaginary rollout stages, agents are allowed to transmit only their stochastic states $z^i_t$. As in DreamerV2, stochastic state has a discrete nature and consists of $32$ categorical distributions with $32$ classes each. Straight-through gradients \cite{bengio2013estimating} are used to learn these discrete representations. By design, stochastic state is the only message that can be  broadcasted by the agent, which amounts only to $32\log{32}=160$ bits of information per message. This is largely because of the sparsity of the $z^i_t$ compared to the other possible distributions. We can also use a language-based interpretation \cite{foerster2016learning, lazaridou2016multi} of $z^i_t$ as we restrict agents to a very limited vocabulary to express their current states. We further elaborate on this point in the next section. 

\paragraph{Communication Block} We use Transformer architecture \cite{vaswani2017attention} with 3 stacked layers for encoding state-action pairs $(\textbf{z}_t, \textbf{a}_t)$. These encodings $e^i_t$ are then used by the agents to update the world model and make action predictions. We use a dropout technique \cite{srivastava2014dropout} with probability $p=0.1$ to prevent the model from overfitting and use positional encoding \cite{vaswani2017attention} to distinguish agents in homogeneous settings. 

\paragraph{Absorbing states.} One major difference between single- and multi-agent environment dynamics is the varying number of agents on the map. Even if the starting number of agents is always the same, it is natural that agents will have varying trajectory lengths depending on the task. Thus, it is essential to handle the states of the agents that are no longer present in the environment during the training phase. Absorbing state \cite{kostrikov2018discriminator} is an imaginary state that does not yield a reward for any action and does not lead to any other state, creating an infinite loop to itself. In our implementation, we add absorbing states after the end of an agents trajectory, so that it does not get a reward for future actions and so other agents can see that that agent is no longer on the map. We do not use observation loss to learn the representation of those states as no meaningful observation can be provided by the environment. Instead, we allow the world model to implicitly learn those representations only from predicting the discounts and rewards that are both equal to zero.

\textbf{Behaviour learning.} We chose Actor-Critic framework to learn an agent's behaviour, where both actor and critic are parameterized by neural networks $\psi$ and $\eta$ respectively:
\begin{align*}
\begin{array}{ll}  
   \text{Actor:}  &a^i_t \sim p_\psi(a^i_t \mid \hat{z}^i_t, h^i_t)\\
   \text{Critic:} &v_\eta^i(\hat{\textbf{z}}_t) \approx \mathop{\mathbb{E}}_{p_\psi}\left[\overset{\infty}{\underset{\tau=t}{\sum}} \gamma^i_\tau r^i_\tau\right]\\
\end{array}
\end{align*}

\paragraph{Imaginary rollouts} As in Dreamer, we use learned transition function $p_\phi(\hat{z}^i_t \mid h^i_t)$ to predict the next stochastic state based on agents' actions and current states of the model. Starting from initial states sampled from the replay buffer, we generate rollouts based on current policies of the agents in order to deploy on-policy learning algorithms in the latent space.  We set the length of the imaginary rollouts to $H=15$ to account for increased complexity of the multi-agent environments and possible compounding error \cite{janner2019trust} of the world model.

\begin{figure}[h]
\centering
\includegraphics[width=0.95\linewidth]{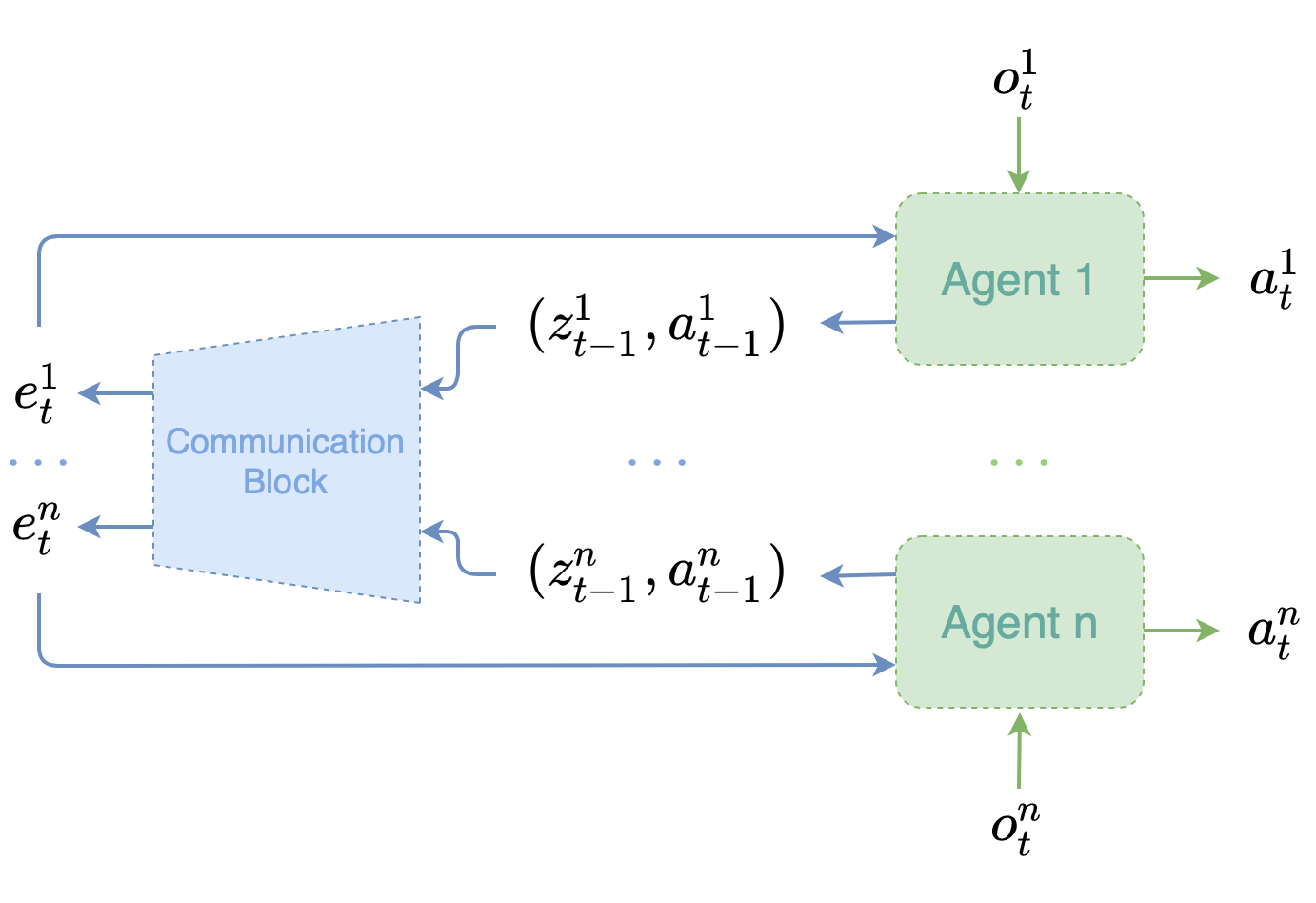}
\caption{Execution Phase. Each agent has its own version of the world model, which can be updated using the Communication Block. It is sufficient to send only stochastic state $z^i_{t-1}$ and action $a^i_{t-1}$ from the previous step for each agent in order to obtain feature vectors $e^i_t$. Agent $i$ then can use the updated world model and its current observation $o^i_t$ to output its next action $a^i_t$.}
\label{fig:scheme_exec}
\end{figure}

\begin{figure*}[ht]
\centering
\begin{subfigure}{.45\linewidth}
\centering
\includegraphics[width=\linewidth]{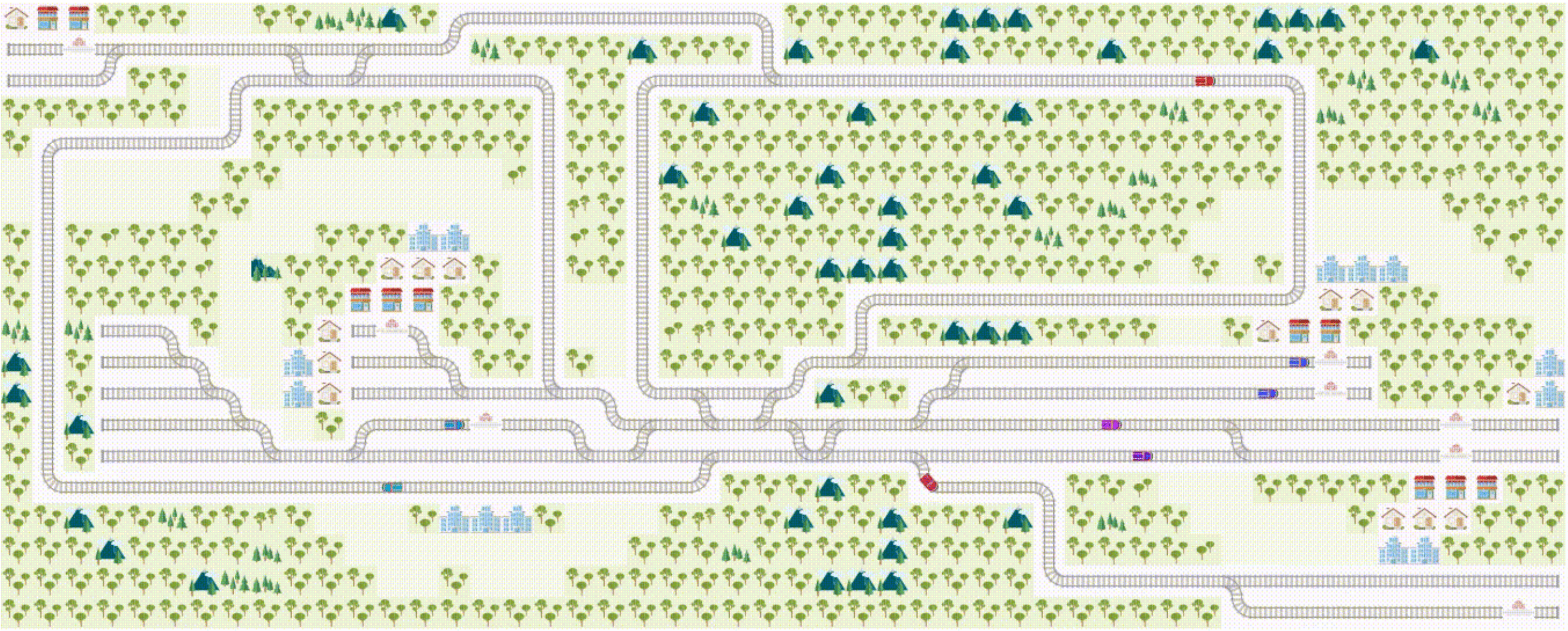}
\caption{Flatland}
\end{subfigure}%
\hfill
\begin{subfigure}{.4\linewidth}
\centering
\includegraphics[width=\linewidth]{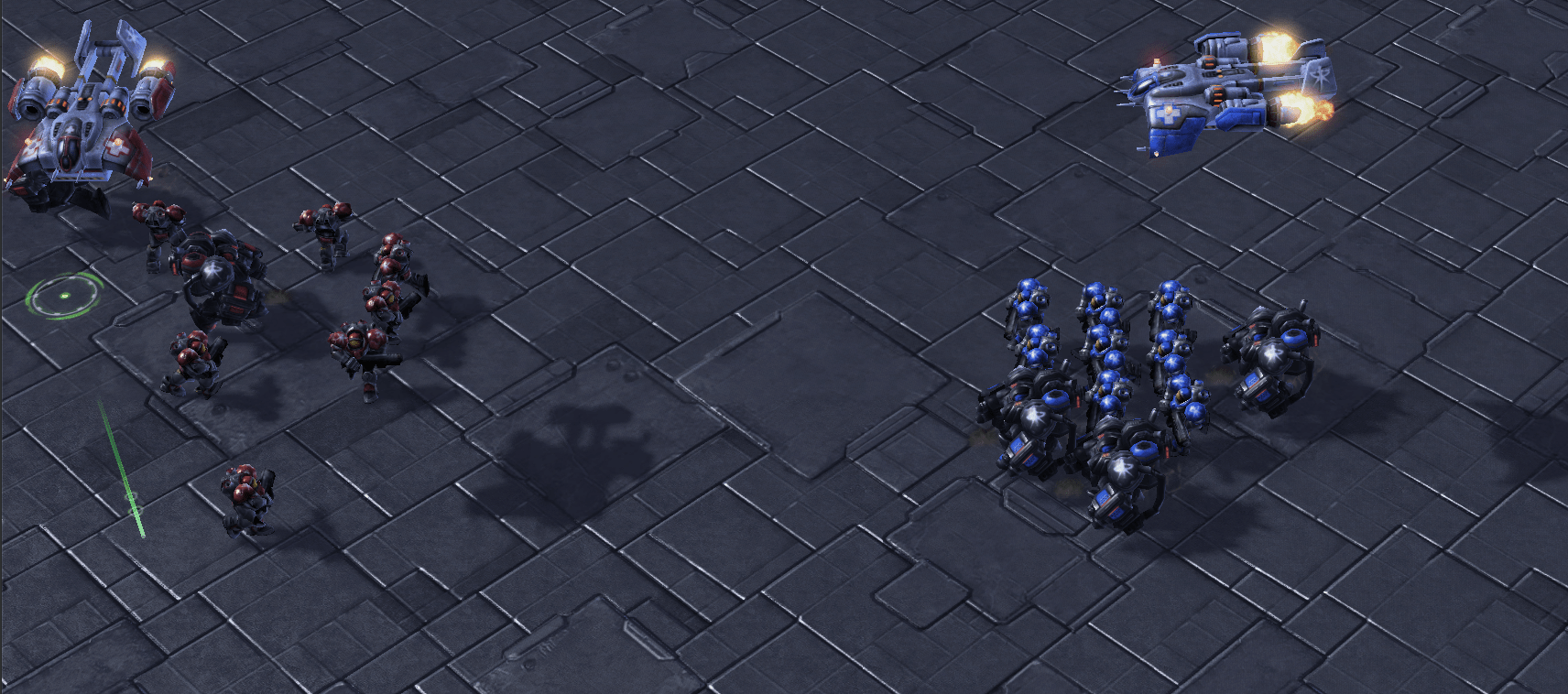}
\caption{StarCraft Multi-Agent Challenge}
\end{subfigure}
\caption{Environments}
\label{fig:env}
\end{figure*}

\paragraph{Actor} Instead of using Reinforce \cite{williams1992simple} for policy updates as in Dreamer, we found that for multi-agent environments PPO updates (Eq. \ref{eq:loss_ppo}) yield better results. We do not use gradients backpropagated through the dynamics model as they hinder the performance, which is likely due to its high noisiness in multi-agent environments. We use parameter sharing \cite{gupta2017cooperative} for the actor network to accelerate training.

\paragraph{Critic}\label{par:critic} We use $\lambda$-target \cite{schulman2015high} for value function as in Dreamer, which gives us more unbiased estimates than TD-error \cite{sutton2018reinforcement}. We augment critic by giving it as an input all current latent states $\hat{\textbf{z}}_t$, which are processed using Attention mechanism. Unlike other CTDE approaches \cite{lowe2017multi, rashid2018qmix}, we cannot use additional information beyond the states predicted by the model as we train agents in the latent space. However, several studies \cite{de2020independent, wang2020off} suggested that such additional information can cause centralized-decentralized mismatch, which can hinder the learning process due to the large variance of the policy gradient updates. Furthermore, in real world scenarios we would not have such additional information as the global environment may be too large. 

\subsection{Communication}

Here, we discuss the properties of our communication protocol. Similar to methods employing Attention mechanism to collect useful information from other agents \cite{liu2020multi, das2019tarmac, jiang2018learning}, we allow agents to broadcast their latent state representations from the previous step $\textbf{z}_{t-1}$. We also assume that agents can observe actions $\textbf{a}_{t-1}$ of others. Tuples $(\textbf{z}_{t-1}, \textbf{a}_{t-1})$ are then transformed by Communication Block to feature vector $e^i_t$ that is used to update the agent's world model and predict next action based on the updated latent state. The schematic representation of the execution phase is in Figure \ref{fig:scheme_exec}.

\paragraph{Reward-agnostic communication} One important distinction of MAMBA communication from other communication protocols \cite{sukhbaatar2016learning, foerster2016learning, das2019tarmac} is that it is not learned from the reward signal. We will refer to our communication protocol as reward-agnostic communication as we use a world model to obtain it, and we will denote communication protocols from \cite{sukhbaatar2016learning, foerster2016learning, das2019tarmac} as goal-oriented communication as they are shaped by the agent's reward. We argue that reward-agnostic communication is more suited for language that describes current environment while goal-oriented communication is more suited for describing agent's task. This duality is mirrored in language theory as representation and acquisition theories respectively \cite{ellis2006language}. We can think of reward-agnostic communication as an emergent language that agents cannot alter to their needs but in exchange this language can provide an unbiased representation of the current state of the agent. On the other hand, goal-oriented communication is defined by the needs of the agent, thus making it a useful tool to achieve its goals. Consequently, many previous works on communication \cite{cao2018emergent, blumenkamp2020emergence, noukhovitch2021emergent} studied this phenomenon as selfish agents tried to manipulate the communication channel to their own advantage by sending false messages to others. Conversely, we cannot use reward-agnostic communication in mixed environments as it provides true information about the agent to both allies and competitors. 

\paragraph{Locality} As we want to preserve locality as much as possible, we can make the agent receive messages only from its neighbours in the environment. Attention mechanism allows us to vary the number of messages and, assuming that agents which are not neighbours will not have an impact on the agent transition function, it is possible to mask weights of those agents in Attention.

\paragraph{Discrete communication} Multiple studies have shown that discrete messages enhance performance as agents develop the language to communicate with each other \cite{havrylov2017emergence, lazaridou2016multi, foerster2016learning}. On the other hand, language capabilities to succinctly describe images in only a few words have inspired several studies in representation learning \cite{oord2017neural, razavi2019generating} to use discrete stochastic states. In our communication protocol, we can see the duality of these ideas as we can train stochastic states on unsupervised loss to reason about the current state of the agent and at the same time use it to communicate agent's current surroundings.

\paragraph{Computation-Communication trade-off} Broadcasting only tuples $(\textbf{z}_{t-1}, \textbf{a}_{t-1})$ to all agents requires $\mathcal{O}(n^2)$ computations on each agent to calculate single-layer attention encodings \cite{vaswani2017attention}. Alternatively, agents can compute only one vector from $n$, for example, $q^iK^T$ from Eq. \ref{eq:attn}, and exchange computed vectors on each step of attention to reduce computation complexity to $\mathcal{O}(n)$.

\subsection{Limitations and Improvements}
MAMBA satisfies all of the stated desiderata by leveraging a Model-Based approach for large scale multi-agent environments. However, MBRL has several limitations compared to MFRL methods. First, learning the world model introduces additional complexity to the Reinforcement Learning algorithm, which requires more careful tuning of hyperparameters. Second, while MAMBA is able to achieve good performance in most environments in a low data regime, it also requires more computational time for one collected trajectory compared to MFRL methods as it needs to learn the world model and rollout imaginary trajectories to learn the policy. We provide detailed comparisons of average wall clock time for training and execution with the same hardware in Appendix. 


\section{Experiments}

\begin{figure*}[ht]
  \centering
\includegraphics[ width=.52\linewidth]{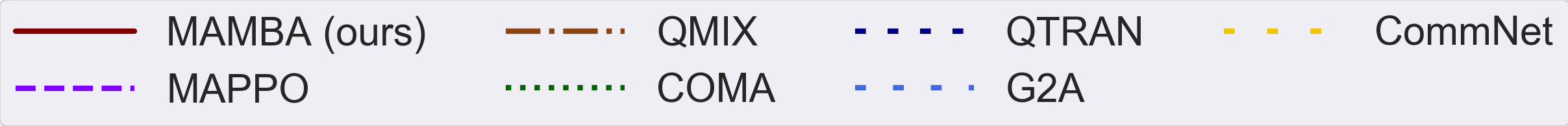}
    \centering
    \begin{subfigure}{.247\textwidth}
      \centering
      \includegraphics[width=\linewidth]{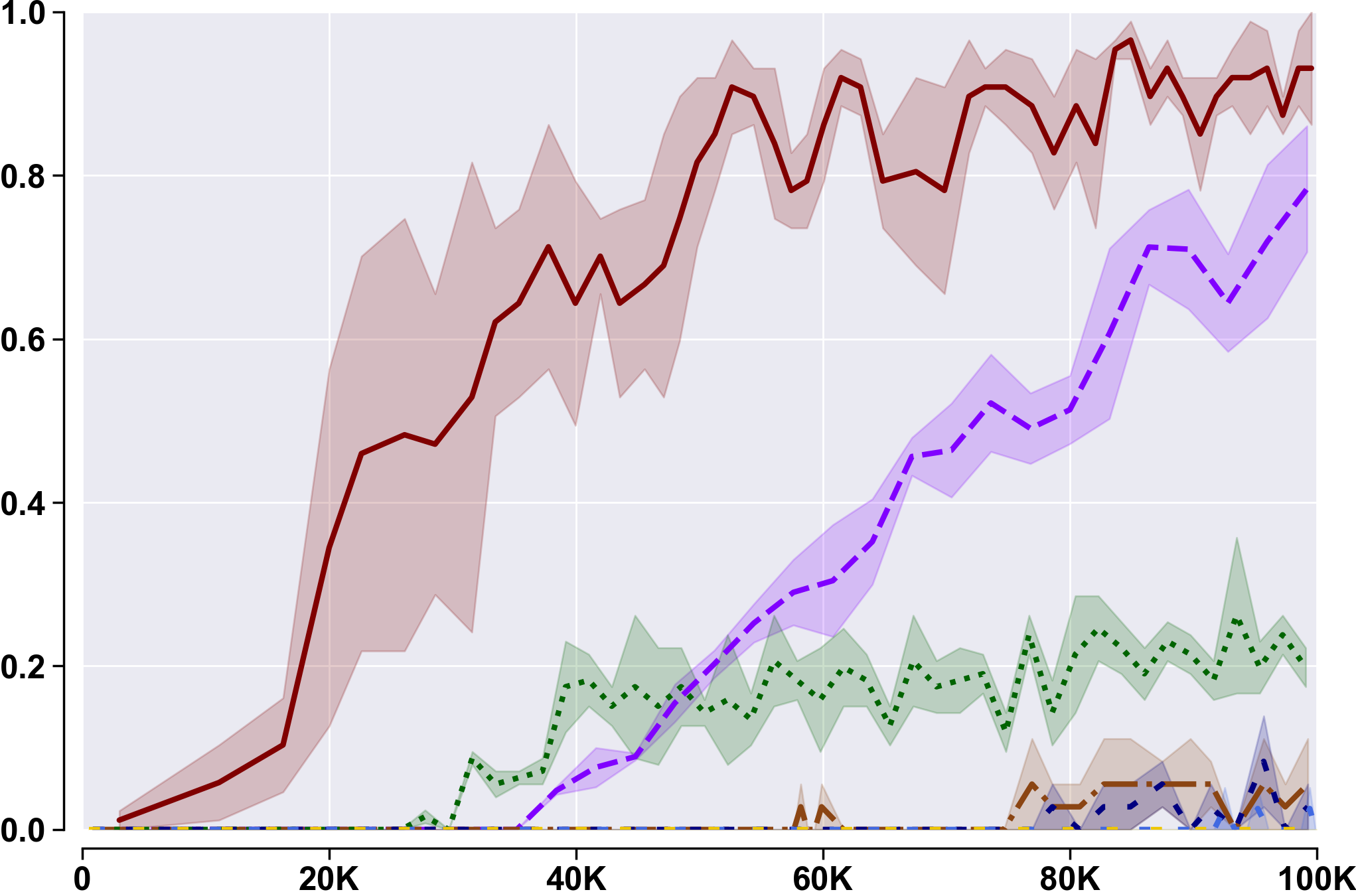}
      \caption{2s vs 1sc}
    \end{subfigure}%
    \begin{subfigure}{.247\textwidth}
      \centering
      \includegraphics[width=\linewidth]{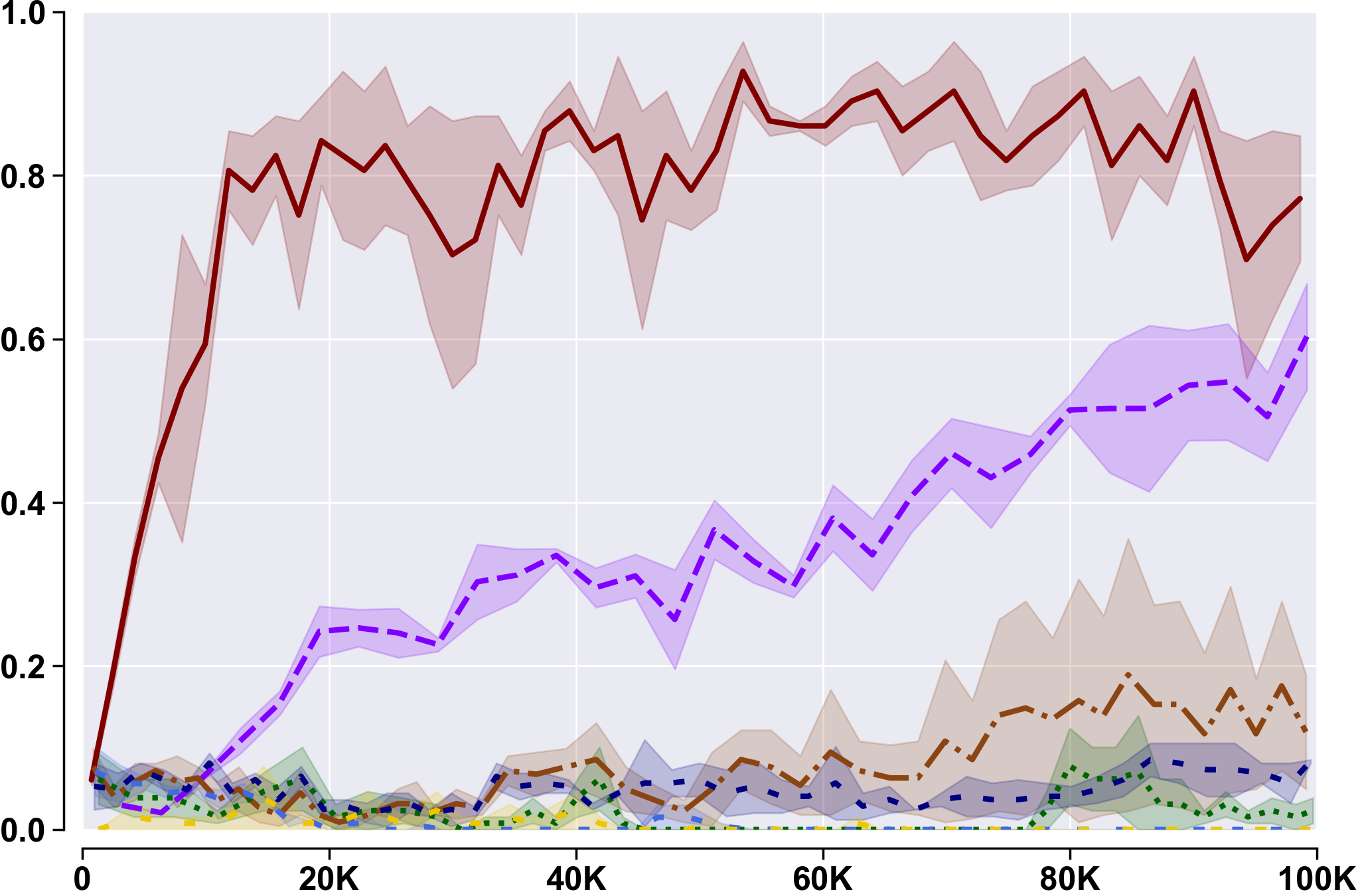}
      \caption{so many baneling}
    \end{subfigure}
    \begin{subfigure}{.247\textwidth}
      \centering
      \includegraphics[width=\linewidth]{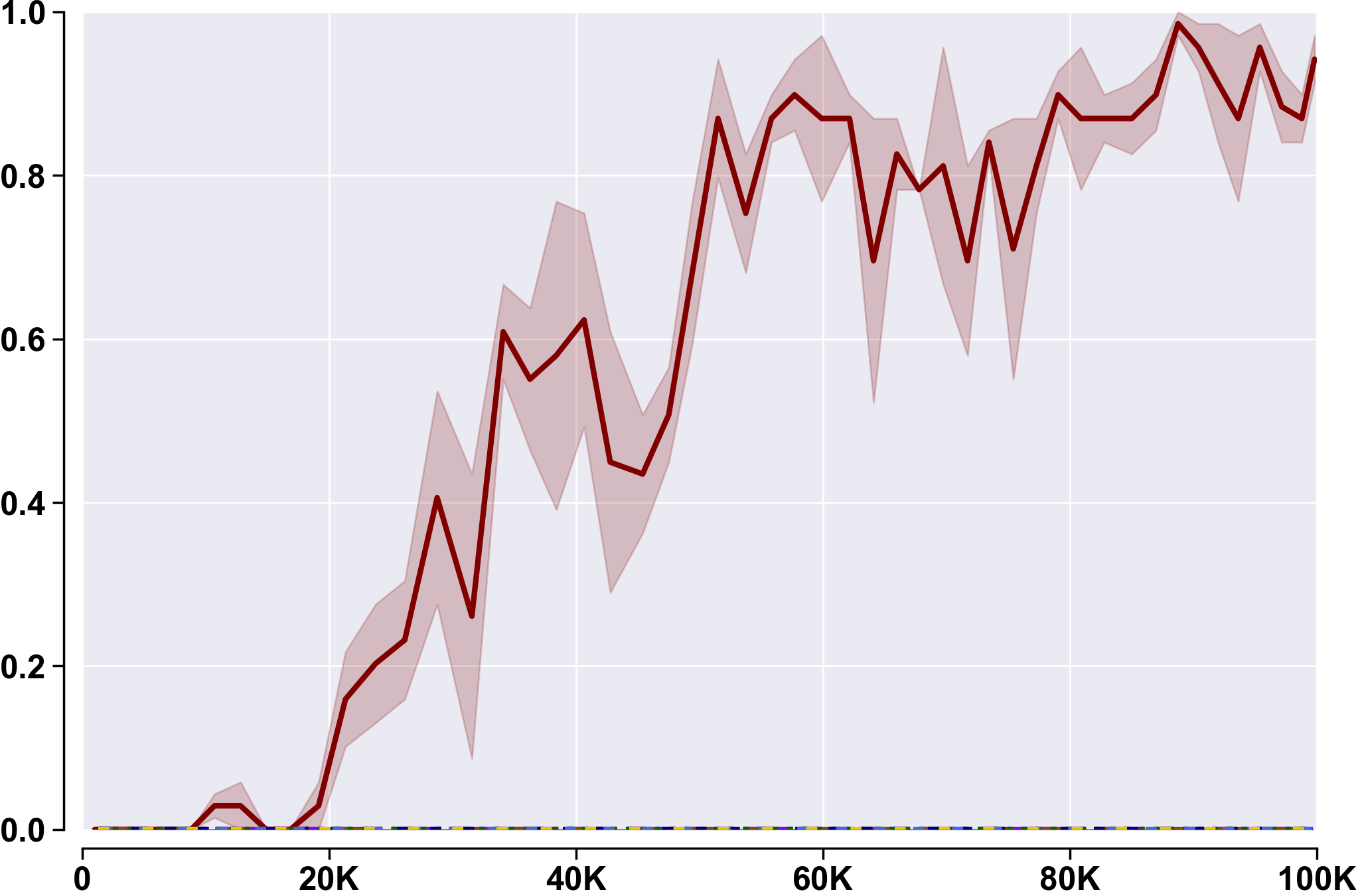}
      \caption{3s vs 3z}
    \end{subfigure}
    \begin{subfigure}{.247\textwidth}
      \centering
      \includegraphics[width=\linewidth]{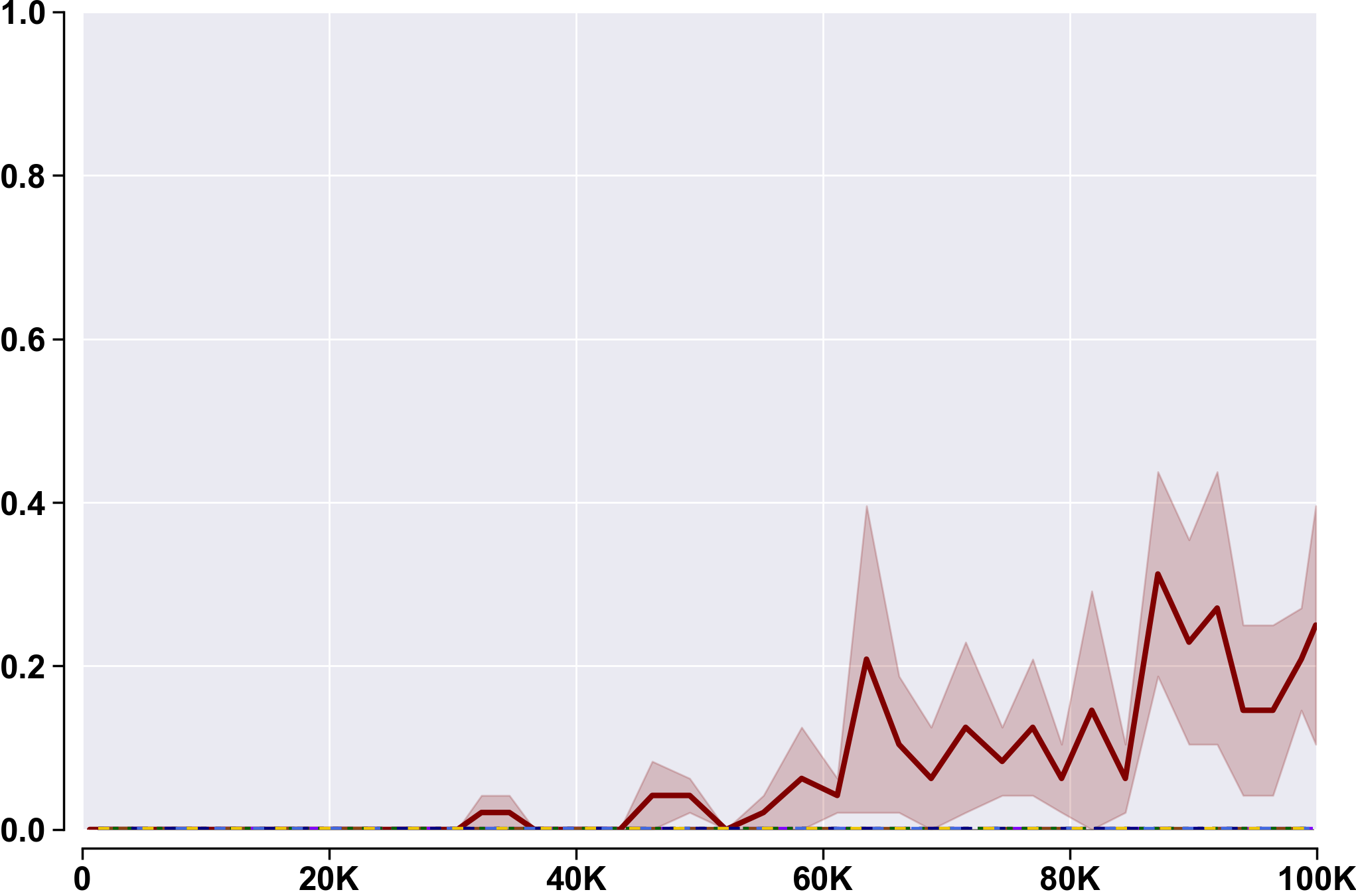}
      \caption{3s vs 4z}
    \end{subfigure}

    \centering
    \begin{subfigure}{.247\textwidth}
      \centering
      \includegraphics[width=\linewidth]{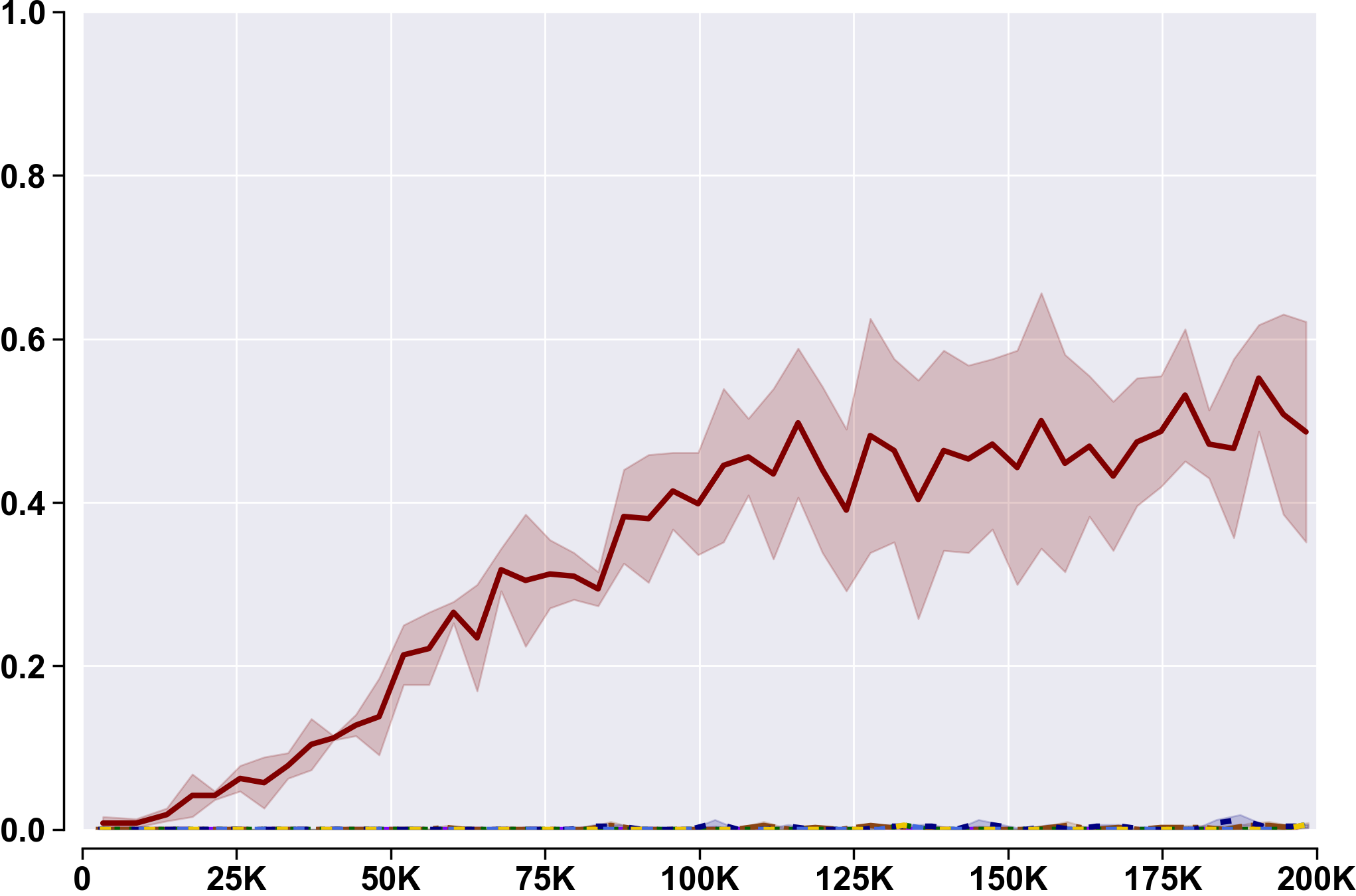}
      \caption{8m vs 9m}
    \end{subfigure}
    \begin{subfigure}{.247\textwidth}
      \centering
      \includegraphics[width=\linewidth]{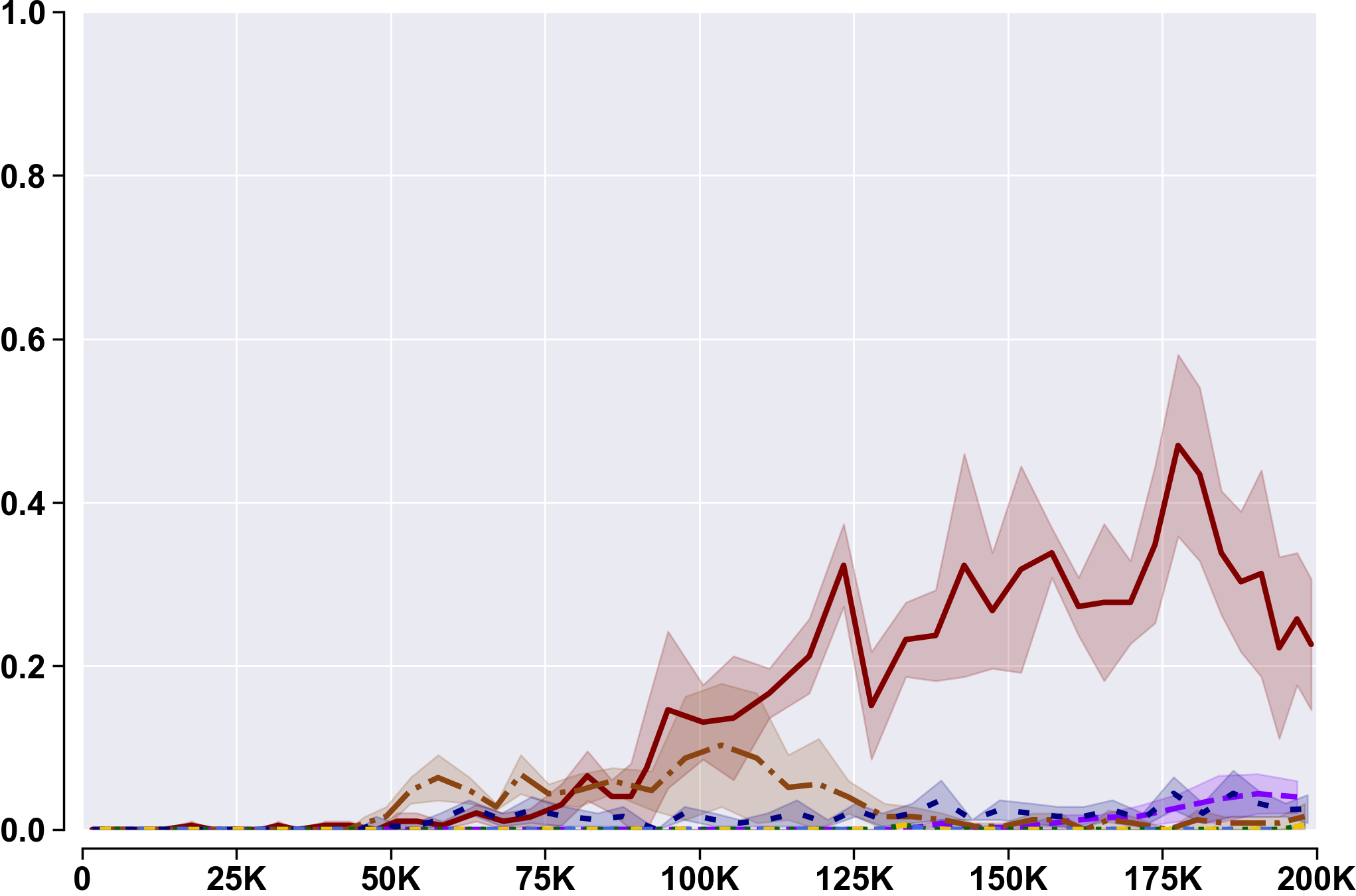}
      \caption{25m}
    \end{subfigure} 
    \begin{subfigure}{.247\textwidth}
      \centering
      \includegraphics[width=\linewidth]{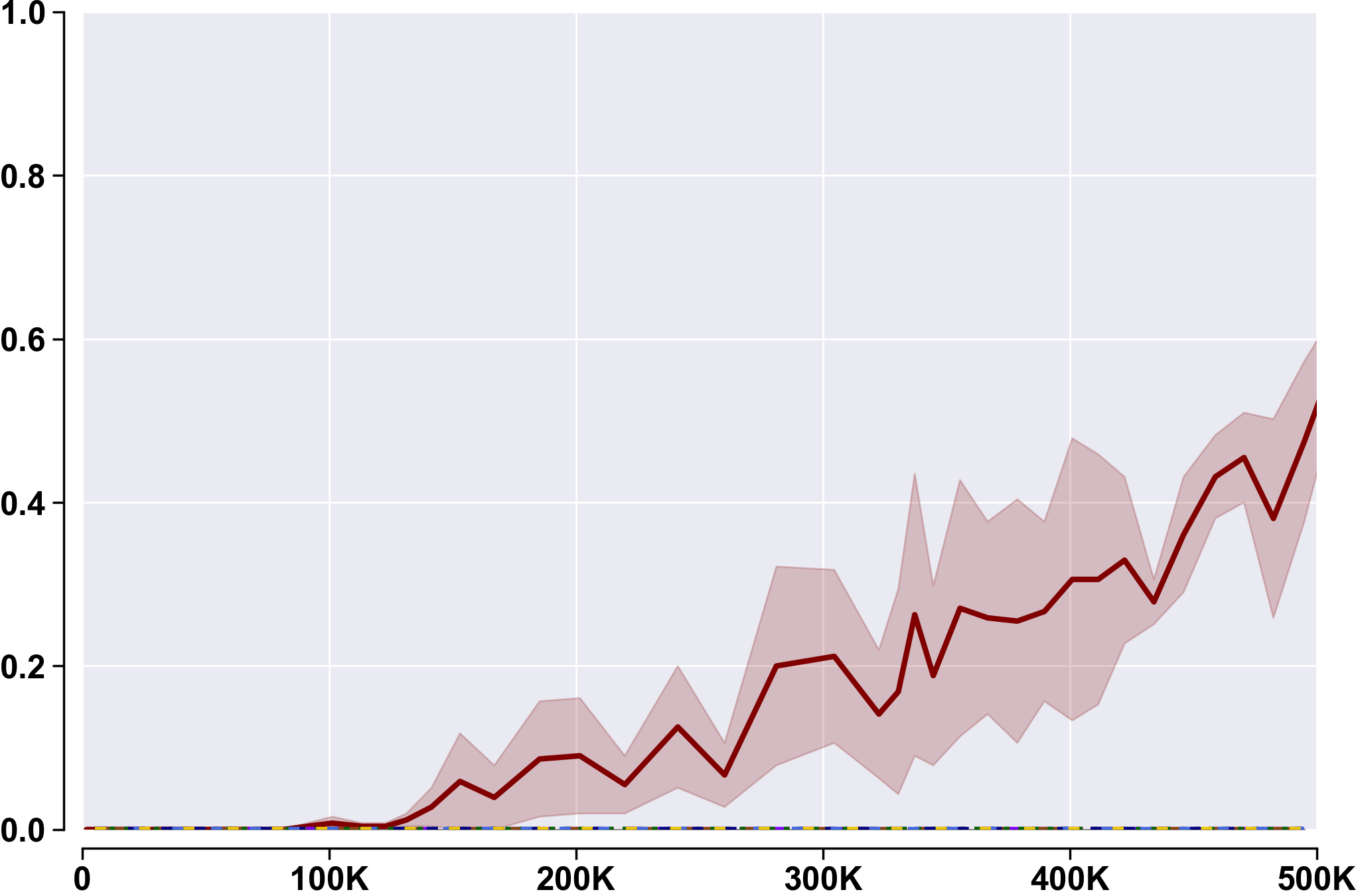}
      \caption{corridor}
    \end{subfigure}%
    \begin{subfigure}{.247\textwidth}
      \centering
      \includegraphics[width=\linewidth]{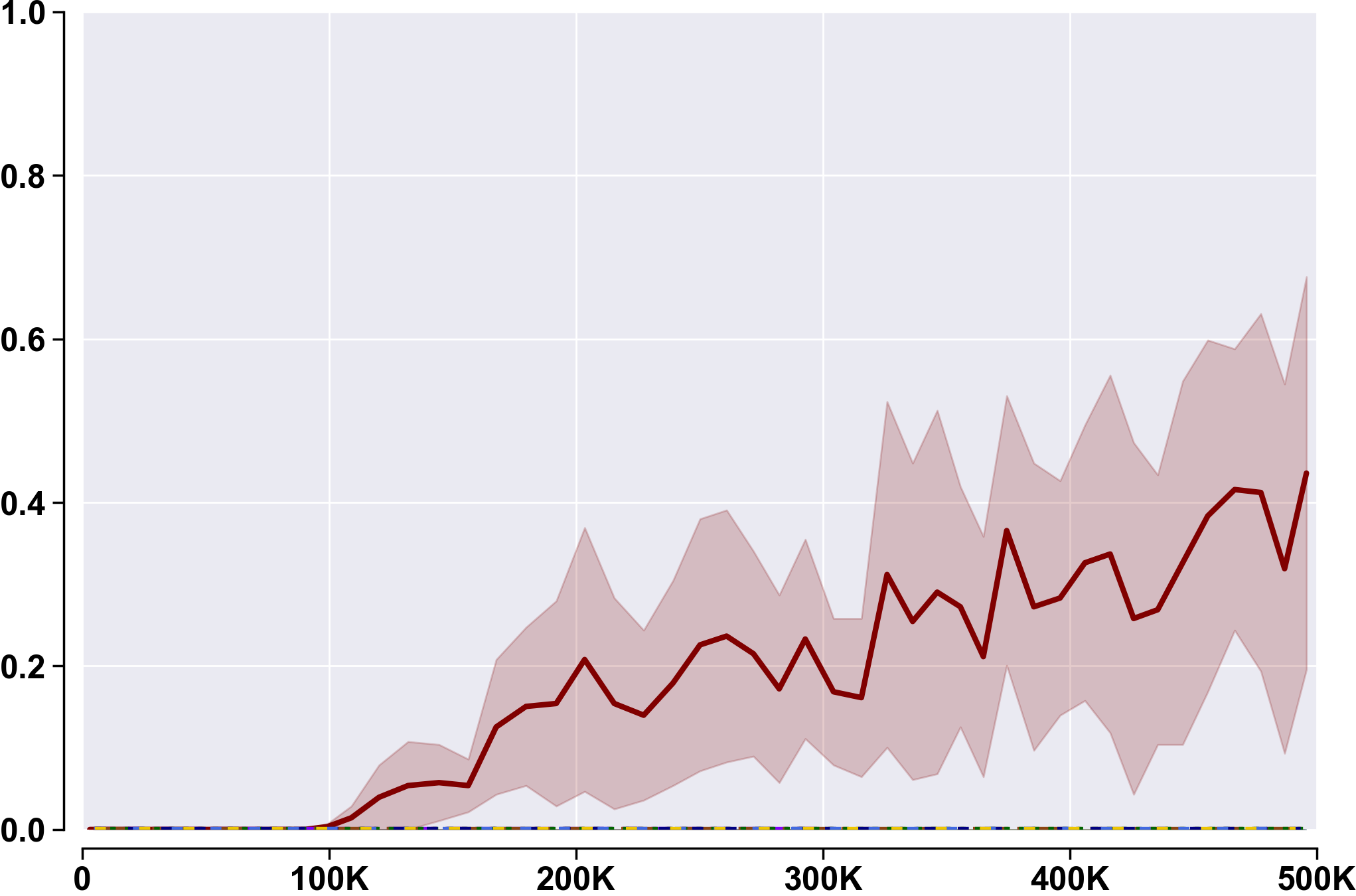}
      \caption{3s5z vs 3s6z}
    \end{subfigure}
    
  \caption{Experiments in SMAC environments. Y axis denotes winning rate of the agents and X axis denotes number of steps taken in the environment.}
  \label{fig:starcraft_exp}
\end{figure*}

\subsection{Environments}
\paragraph{Flatland} \cite{mohanty2020flatland} (Figure \ref{fig:env}a) is a multi-agent 2d environment simulating train traffic on the rail network. Each agent has its own starting point and destination and the goal of the environment is to guide agents to their destinations without collisions. The difficulty of the environment can be flexibly adjusted as the number of agents as well as the size of the map could be easily configured. Note that maps are generated procedurally for each trajectory, so the world model must be able to generalize well in order to achieve good performance. Each agent receives positive reward on arrival to its destination and negative reward if a collision occurs. We additionally use more dense variant of the reward provided by the winning solution of Flatland 2020 RL challenge \cite{laurent2021flatland}. Agent receives additional positive reward if its action shortens the approximated distance to its destination. This reward allows us to alleviate credit assignment issue and facilitates exploration for better performance. In this environment, the density of the traffic can be quite sparse, allowing us to test locality of our approach. During the execution phase, the agents can send messages only to their neighbours, which are defined by their proximity on the railroad.

\paragraph{StarCraft Multi-Agent Challenge} (SMAC) \cite{samvelyan19smac} (Figure \ref{fig:env}b) is a suite of cooperative multi-agent environments based on StarCraft II. Each task consists of a single scenario with two confronting teams, one of which is controlled by the game bot and the other by our algorithm. The goal of the environment is therefore to defeat enemy team. SMAC offers 23 different scenarios with varying difficulty and number of agents. SMAC focuses on learning micromanagement of a group of agents, e.g. using terrain to your advantage as in corridor map or kite opponents as in 3s vs 3z map. Learning this behaviours requires both good exploration properties as well as solving credit assignment issue as agents need to disentangle global reward for their actions. One important property of the environment is that not all actions are available during agents decision making, e.g. attacking opponents that are out of reach. This necessitates world model to predict if the action is available or not based on latent state in order to train agents in imaginary rollouts. We use additional Bernoulli classifier similar to Discount predictor and train it on action masks provided by the environment for each agent.

\subsection{Results}
We adopt low data training regime from \cite{kaiser2019model} to highlight sample efficiency of the current state-of-the-art MFRL methods. We chose MAPPO \cite{yu2021surprising}, COMA \cite{foerster2018counterfactual}, QMIX \cite{rashid2018qmix}, CommNet \cite{sukhbaatar2016learning} and G2A \cite{liu2020multi} as the baselines for SMAC environments and winning solution of \cite{laurent2021flatland} as the baseline for Flatland environment. There are three levels of difficulty both in SMAC and Flatland. In order to make up for this difference, we allocate more samples from the environment depending on its difficulty. Specifically, we use 100k, 200k and 500k samples for SMAC and 300k, 1kk and 1.5kk for Flatland. All hyperparameters, ablation study and details about the environments can be found in Appendix. We use this implementation\footnote{\href{https://github.com/starry-sky6688/StarCraft}{https://github.com/starry-sky6688/StarCraft}} for all baselines in SMAC except MAPPO, which implementation is provided by authors\footnote{\href{https://github.com/marlbenchmark/on-policy}{https://github.com/marlbenchmark/on-policy}}. We use authors implementation of the winning Flatland 2020 solution that can be found here\footnote{\href{https://github.com/jbr-ai-labs/NeurIPS2020-Flatland-Competition-Solution}{https://github.com/jbr-ai-labs/NeurIPS2020-Flatland-Competition-Solution}}.

\begin{figure*}[t]
    \centering
    \begin{subfigure}{.33\textwidth}
      \centering
      \includegraphics[width=\linewidth]{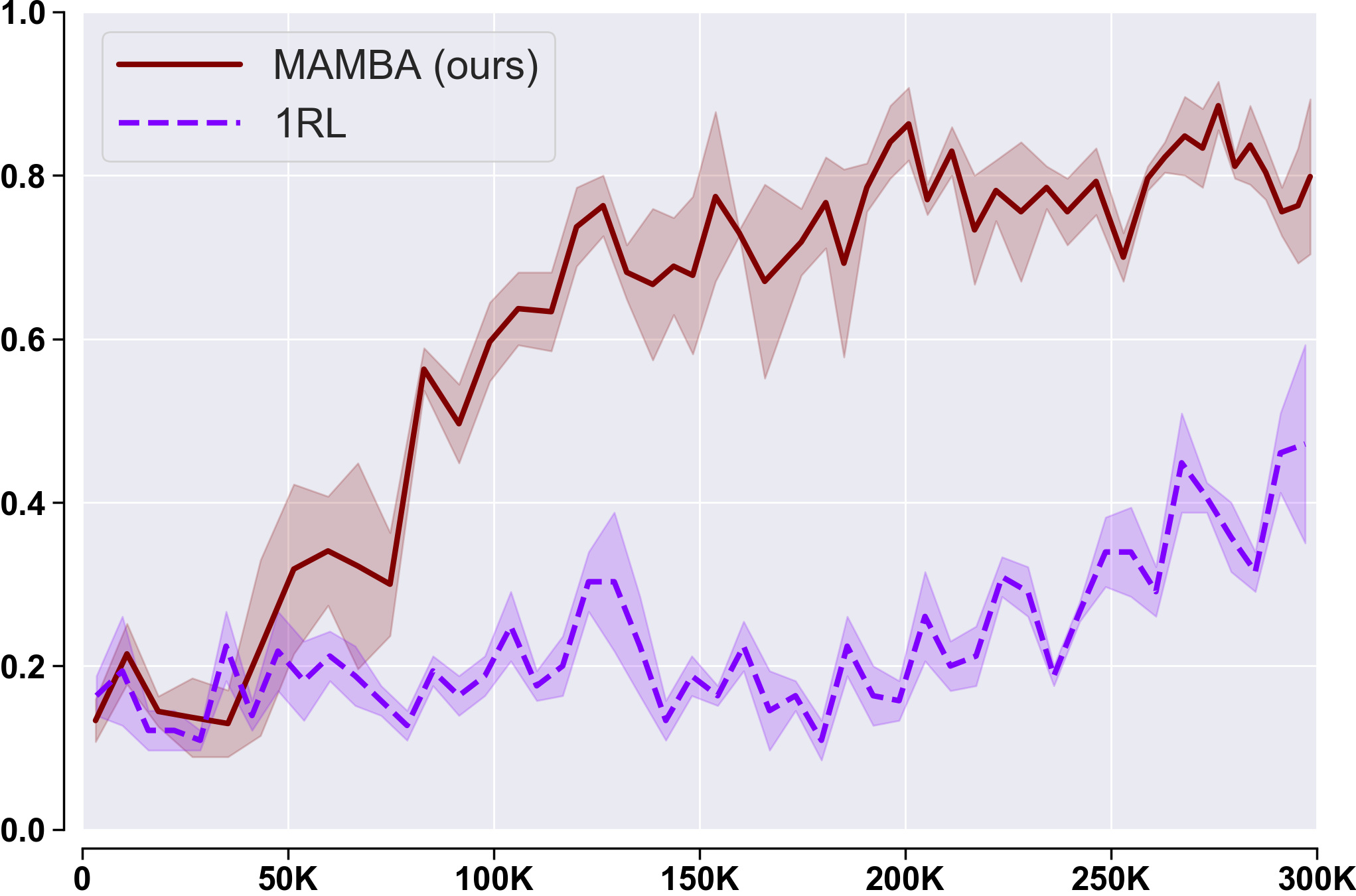}
      \caption{5 agents}
    \end{subfigure}%
    \begin{subfigure}{.33\textwidth}
      \centering
      \includegraphics[width=\linewidth]{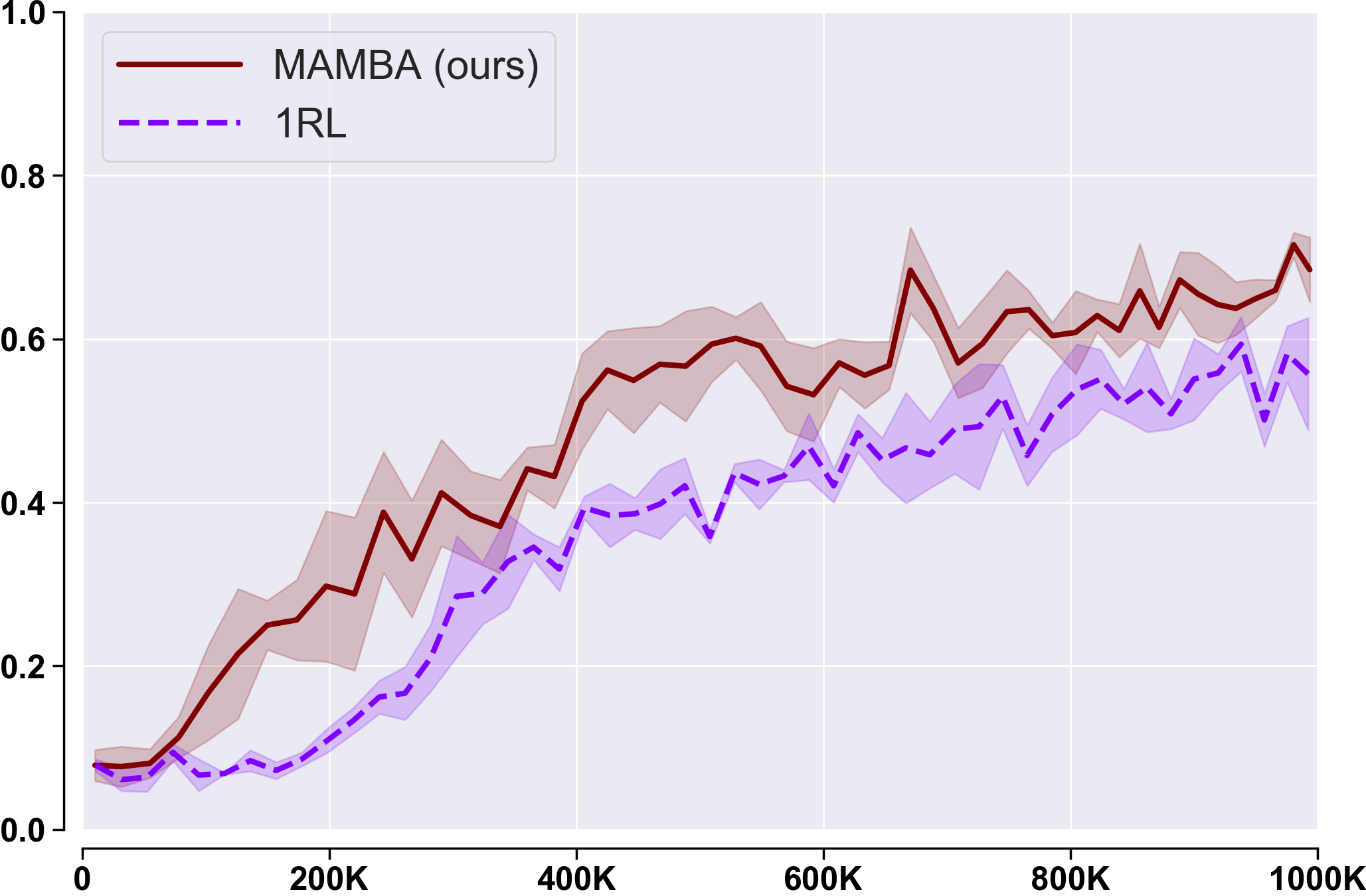}
      \caption{10 agents}
    \end{subfigure}
    \begin{subfigure}{.33\textwidth}
      \centering
      \includegraphics[width=\linewidth]{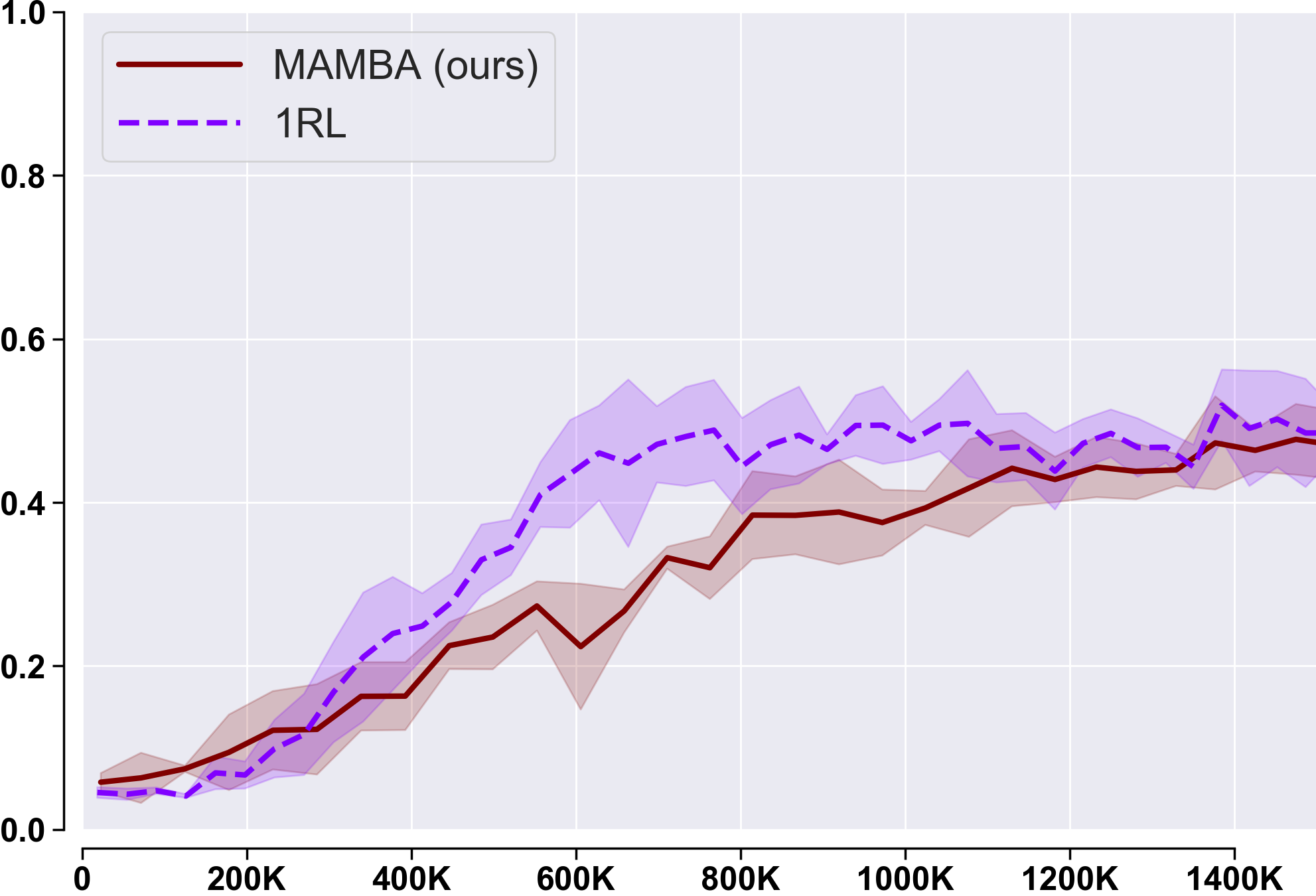}
      \caption{15 agents}
    \end{subfigure}
    
  \caption{Experiments in Flatland environment. Y axis denotes percentage of arrived trains and X axis denotes number of steps taken in the environment.}
  \label{fig:sflatland_exp}
\end{figure*}

\paragraph{SMAC} The results in SMAC environments are presented in Figure \ref{fig:starcraft_exp}. We selected 8 environments out of 23 to show in the paper and the rest of the results can be found in Appendix. Specifically, we chose 4 random environments (Fig. \ref{fig:starcraft_exp} a, b, c, d) that are considered easy tasks, 2 environments (Fig. \ref{fig:starcraft_exp} e, f) with large number of agents and 2 environments (Fig. \ref{fig:starcraft_exp} g, h) with very hard tasks that require coordination. SMAC environments require coordination between all agents to complete the task, thus we did not test the locality property of our algorithm here. We can see that MAMBA outperforms all baselines in all of the environments with given number of samples. Notably, MAMBA achieves near-optimal performance on easy tasks with only 100k samples and finds winning strategy for hard and very hard tasks, winning nearly half of the episodes. On the other hand, the current state-of-the-art MFRL method MAPPO, which is able to solve all SMAC environments given large amount of samples, struggles to achieve good performance in low data regime. This highlights sample inefficiency of current MFRL approaches and shows the potential for a significant reduction in needed samples using MBRL methods. Note that communication methods G2A and CommNet are less sample efficient than MAPPO,  which is likely due to additional complexity of learning how to leverage message channel. On the other hand, MAMBA encapsulates communication in a world model, which can be learned in fewer samples. Two important highlights of SMAC environments are the credit assignment problem and usage of additional information during training. We can see that MAMBA is able to disentangle global rewards and achieve good performance even without additional credit assignment techniques in most environments. We do not use additional global information during training, which is a limitation of model-based approaches discussed in Section \ref{par:critic}. 

\paragraph{Flatland} The results in Flatland environment are presented in Figure \ref{fig:sflatland_exp}. We use modified version of the winning solution from the RL track of Flatland 2020 challenge \cite{laurent2021flatland} as our main baseline, which we will refer to as 1RL. Our only modification is disabling departure schedule, as it alters the underlying MDP of the environment. As there is no global reward in the environment, we do not compare with algorithms specific for solving credit assignment issue, e.g. COMA, QMIX and QTRAN. We also found it redundant to compare with MADDPG or communication methods such as G2A or CommNet as the winning solution employs PPO-based method with both augmented critics and communication between agents. In this environment, we use three different maps with the increasing size and number of agents. That way, we can progressively increase the difficulty of the environment and see how this affects the algorithms. Specifically, we use environments with 5, 10 and 15 agents, which turn traffic on the rail network more and more dense. This increase in difficulty affects the learning of world model as it must predict next states of all agents during training. Furthermore, during execution we provide agent messages only from its neighbours for both 1RL and MAMBA, thus limiting information about the environment only to local area of the map. Nevertheless, we can see that MAMBA can successfully achieve good performance in the 5 agents (Fig. \ref{fig:sflatland_exp} a) setting only in 300k samples while 1RL still needs more samples to improve performance. Naturally, performance decrease in 10 agents setting for both algorithms, but we can see that MAMBA still substantially outperforms 1RL method in 1 million samples (Fig. \ref{fig:sflatland_exp} b). However, we can see that sample efficiency starts to equalize compared to 5 agents setting. We hypothesize that it is easier for 1RL agents to learn on raw observations with communication than for MAMBA agents to learn from more noisy discrete latent space of the model. This issue of decaying sample efficiency is further exacerbated with the increased number of agents in the next setting. In 15 agents setting (Fig. \ref{fig:sflatland_exp} c) we can see that MAMBA is less sample efficient than 1RL but still achieves comparable performance in 1.5 million samples. We hypothesize that this decrease in sample efficiency results from the increased difficulty of the environment, which necessitates more expressive world model to approximate as well as more samples to learn. 

\section{Conclusion}
In this paper, we present MAMBA -- a MBRL method for multi-agent cooperative environments. We show the effectiveness of MAMBA in low data regime on two challenging domains of SMAC and Flatland compared to current state-of-the-art approaches, highlighting sample inefficiency of the MFRL methods. We provide architecture that can scale to a large number of agents as well as a training procedure that can learn disentangled latent space for agents. This allows us to decompose the world model by the agent, which can then be used in a decentralized manner with communication. Furthermore, we use discrete messages to facilitate decentralization and account for message channel bandwidth limitations. 

\section{Acknowledgements}

The publication was supported by the grant for research centers in the field of AI provided by the Analytical Center for the Government of the Russian Federation (ACRF) in accordance with the agreement on the provision of subsidies (identifier of the agreement 000000D730321P5Q0002) and the agreement with HSE University  No. 70-2021-00139. This research was supported in part through computational resources of HPC facilities at HSE University \cite{kostenetskiy2021hpc}.

\balance

\bibliographystyle{ACM-Reference-Format}
\bibliography{main}

\newpage

\appendix

\section{Hyperparameters}
 We use random search with 50 runs to find best hyperparameters on the Flatland environment with 5 agents and use them for all Flatland environments. Similarly, we use random search with 50 runs for SMAC on 3s vs 3z environment. Specifically, we sample values for search from the following grid: Actor Learning rate from [1e-4, 5e-4, 2e-4], Critic Learning rate from [1e-4, 5e-4, 2e-4], Model Learning rate from [1e-4, 2e-4, 1e-3], Model number of epochs from [20, 40, 60], PPO epochs per update $[3, 5, 10]$, Rollout horizon from [5, 10, 15].
\FloatBarrier
\begin{table}[h!]
\centering
\begin{tabular}{lccc}
\hline
Hyperparameter                        &  Flatland & SMAC\\ \hline
                                      &  &\\
\hspace{3mm}\textbf{PPO}              &  &\\
\hspace{3mm}Batch size                      & \multicolumn{2}{c}{2000}  \\
\hspace{3mm}GAE $\lambda$                   & \multicolumn{2}{c}{0.95 }  \\
\hspace{3mm}Entropy coefficient             & \multicolumn{2}{c}{0.001}    \\
\hspace{3mm}Entropy annealing               & \multicolumn{2}{c}{0.99998}   \\
\hspace{3mm}Number of updates               & \multicolumn{2}{c}{4}  \\
\hspace{3mm}Epochs per update               & \multicolumn{2}{c}{5}    \\
\hspace{3mm}Update clipping parameter       & \multicolumn{2}{c}{0.2} \\
\hspace{3mm}Actor Learning rate             & \multicolumn{2}{c}{5e-4 } \\
\hspace{3mm}Critic Learning rate            & \multicolumn{2}{c}{5e-4 } \\
\hspace{3mm}$\gamma$                        & \multicolumn{2}{c}{0.99}  \\
                              &  \\
\hspace{3mm}\textbf{Model}              &    \\
\hspace{3mm}Model Learning rate               & \multicolumn{2}{c}{ 2e-4 } \\
\hspace{3mm}Number of epochs                  & 40  & 60 \\
\hspace{3mm}Number of sampled rollouts        & \multicolumn{2}{c}{40}   \\
\hspace{3mm}Sequence length                   & 50  & 20  \\
\hspace{3mm}Rollout horizon $H$               & \multicolumn{2}{c}{15}    \\
\hspace{3mm}Buffer size                       & 5e5 & 2.5e5\\
\hspace{3mm}Number of categoricals            & \multicolumn{2}{c}{32}   \\
\hspace{3mm}Number of classes                 & \multicolumn{2}{c}{32}   \\
\hspace{3mm}KL balancing entropy weight       & \multicolumn{2}{c}{0.2}  \\
\hspace{3mm}KL balancing cross entropy weight & \multicolumn{2}{c}{0.8}  \\
                              &  \\
\hspace{3mm}\textbf{Common}                  &    & \\
\hspace{3mm}Gradient clipping                & \multicolumn{2}{c}{100} \\
\hspace{3mm}Trajectories between updates     & \multicolumn{2}{c}{1} \\
\hspace{3mm}Hidden size                      & 400 & 256  \\
\end{tabular}
\caption{Hyperparameters for MAMBA.}
\label{tab:hyperparameters}
\end{table}
\FloatBarrier

\newpage

\section{Environments}

We use authors implementation\footnote{\href{https://github.com/oxwhirl/smac}{https://github.com/oxwhirl/smac}} of SMAC environments with default settings. For Flatland environments, we use Flatland class \cite{mohanty2020flatland} to generate environments with parameters shown in Table \ref{tab:fl}

\FloatBarrier
\begin{table}[h!]
\centering
\begin{tabular}{lccc}
\hline
Flatland parameter                   &  \textbf{5 agents} & \textbf{10 agents} & \textbf{15 agents}\\ \hline
                                      &  &&\\
\hspace{3mm} height                      & 35  & 35  & 40  \\
\hspace{3mm} width                 & 35    & 35  & 45  \\
\hspace{3mm} n\_agents            & 5   & 10  & 15  \\
\hspace{3mm} n\_cities               & 3    & 4  & 5  \\
\hspace{3mm} grid\_distribution\_of\_cities               & False  & False   & False    \\
\hspace{3mm} max\_rails\_between\_cities       & 2   & 2  & 2  \\
\hspace{3mm} max\_rail\_in\_cities                   & 4 & 4 & 4  \\
\hspace{3mm} malfunction\_rate             & 1./100 & 1./150  & 1./200  \\
                              &  &&\\
\end{tabular}
\caption{Parameters for Flatland environments}
\label{tab:fl}
\end{table}
\FloatBarrier

\section{Ablation study}
\begin{figure}[h]
\centering
\centering
\includegraphics[width=\linewidth]{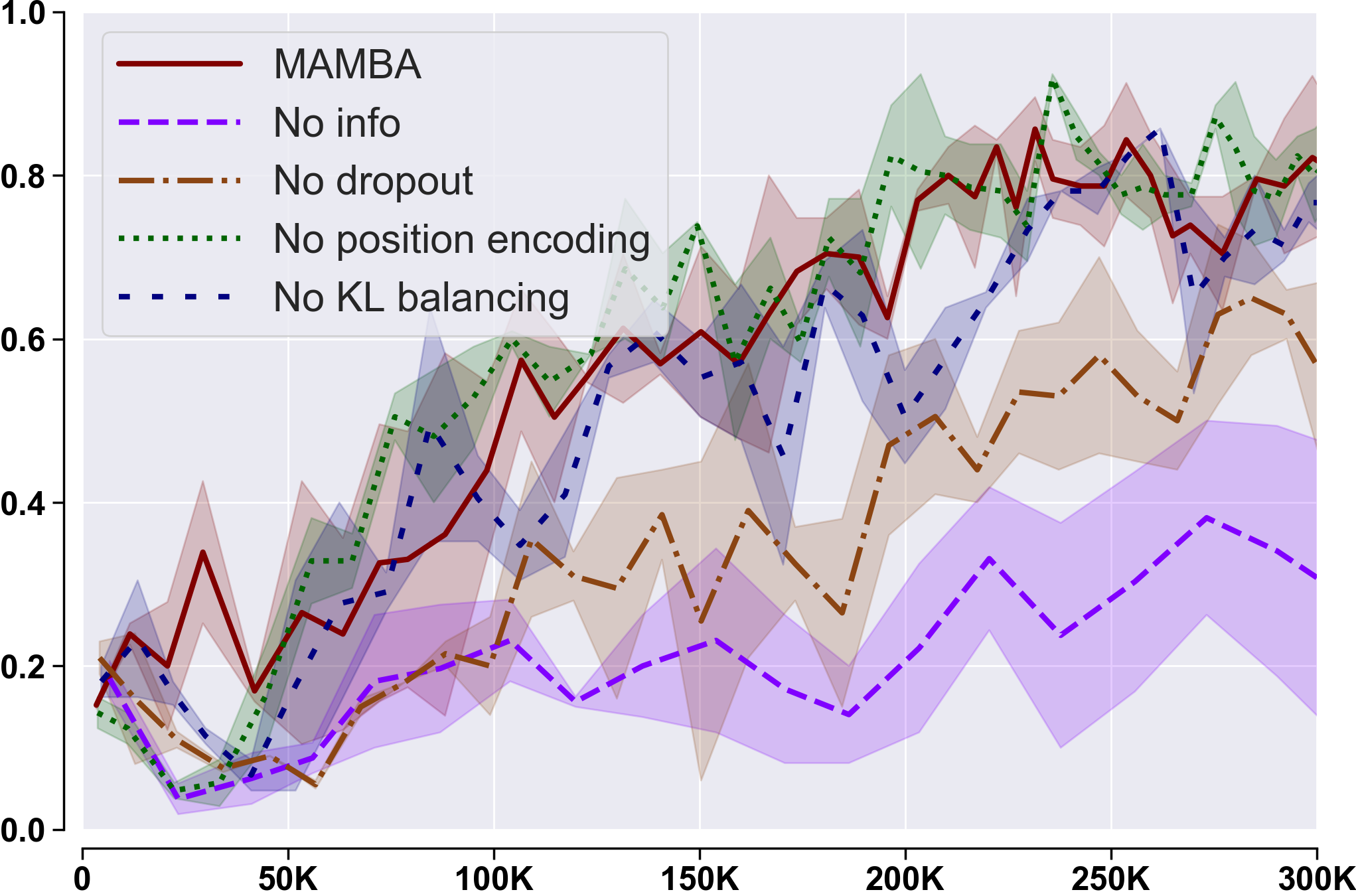}
\caption{Ablation study in Flatland}
\label{fig:abl}
\end{figure}

We conduct several ablation experiments in Flatland environment with 5 agents. The results with algorithms executed with 3 random seeds are reported in Figure \ref{fig:abl}. Specifically, we test whether disabling the information loss $\mathcal{L}_{\text{info}}$, KL balancing, dropout rate $p$ in Communication Block or position encoding in Communication Block degrades final performance. Notably, both position encoding and KL balancing are not important for overall performance while dropout and $\mathcal{L}_{\text{info}}$ have a great impact on learning.

\section{SMAC results}

We report final performance of the algorithms for all SMAC environments in Table \ref{tab:smac}. 

\begin{table*}[h]
\centering
\begin{tabular}{lcccccccc}
\hline
Map                             &  Number of samples & MAMBA                 & MAPPO  & QMIX    & COMA    & QTRAN    & G2A       & CommNet \\ \hline
                                &                    &                       &        &         &         &          &           &         \\
\hspace{3mm} 2m\_vs\_1z         &    100k            &  \textbf{93}(1)       & 89(4)  & 57(12)  &  0      & 57(1)    &  0        &  0       \\
\hspace{3mm} 3m                 &    100k            &  \textbf{91}(2)       & 64(3)  & 53(7)   &  45(6)  & 39(4)    &  8(2)     &   14(4) \\
\hspace{3mm} 2s\_vs\_1sc        &    100k            &  \textbf{98}(1)       & 78(9)  & 10(9)   &  18(2)  &  4(3)    & 11(6)     &    1(2)   \\
\hspace{3mm} 2s3z               &    100k            & \textbf{69}(1) & 3 & 24(4) & 0 & 16(7) & 1(2) & 1 \\
\hspace{3mm} 3s\_vs\_3z         &    100k            & \textbf{90} & 0 & 0 & 0 & 0 & 0 & 0 \\
\hspace{3mm} 3s\_vs\_4z         &    100k            & \textbf{26}(4) & 0 & 0 & 0 & 0 & 0 & 0 \\ 
\hspace{3mm} so\_many\_baneling &    100k            & \textbf{83}(7) & 60(8) & 23(9) & 17(4) & 7(1) & 0 & 0      \\
\hspace{3mm} 8m                 &    100k            & \textbf{80}(3) & 45(7) & 22(11) & 21(1) & 33(6) & 9(1) & 2       \\
\hspace{3mm} MMM                &    100k            & \textbf{48}(11) & 2(1) & 4(3) & 0 & 0 & 0 & 0      \\
\hspace{3mm} 1c3s5z             &    100k            & \textbf{85}(8) & 3(2) & 32(11) & 0 & 5(4) & 0 & 0       \\
\hspace{3mm} bane\_vs\_bane     &    100k            & 39(20) & \textbf{93}(2) & 52(12) & 48(13) & 54(9) & 56(11) & 28(40)       \\
\hspace{3mm} 3s\_vs\_5z         &    200k            & \textbf{3}(4) & 0 & 0 & 0 & 0 & 0 & 0      \\
\hspace{3mm} 2c\_vs\_64zg       &    200k            & \textbf{8}(4) & 0 & 0 & 1(2) & 0 & 0 & 0      \\
\hspace{3mm} 8m\_vs\_9m         &    200k            & \textbf{50}(14) & 0 & 0 & 0 & 0 & 0 & 0     \\
\hspace{3mm} 25m                &    200k            & \textbf{27}(8) & 2(2) & 8(8) & 0 & 3(1) & 0 & 0     \\
\hspace{3mm} 5m\_vs\_6m         &    200k            & \textbf{24}(10) & 0 & 0 & 0 & 0 & 0 & 0      \\
\hspace{3mm} 3s5z               &    200k            & \textbf{17}(5) & 0 & 0 & 0 & 0 & 0 & 0     \\
\hspace{3mm} 10m\_vs\_11m       &    200k            & \textbf{54}(5) & 2 & 1(1) & 0 & 0 & 0 & 0      \\
\hspace{3mm} MMM2               &    500k            & \textbf{23}(3) & 0 & 0 & 0 & 0 & 0 & 0    \\
\hspace{3mm} 3s5z\_vs\_3s6z     &    500k            & \textbf{44}(25) & 0 & 0 & 0 & 0 & 0 & 0      \\
\hspace{3mm} 27m\_vs\_30m       &    500k            &  \textbf{1}(1)      & 0 & 0 & 0 & 0 & 0 & 0     \\
\hspace{3mm} 6h\_vs\_8z         &    500k            & \textbf{1}(2) & 0 & 0 & 0 & 0 & 0 & 0    \\
\hspace{3mm} corridor           &    500k            & \textbf{39}(10) & 0 & 0 & 0 & 0 & 0 & 0    \\
                                &                &           &        &       &       &        &      &         \\
\end{tabular}
\caption{Mean and std of winning percentage after learning for specified number of samples for 3 runs in SMAC environments.}
\label{tab:smac}
\end{table*}

\end{document}